\RequirePackage{snapshot}

\documentclass{article} 
\usepackage{iclr2016_conference,times}

\usepackage{multirow}
\usepackage{longtable}
\usepackage{graphicx}
\usepackage{array}
\usepackage{url}
\usepackage{booktabs}
\usepackage{float}

\title{PixColor: Pixel Recursive Colorization}
\date{May 2017}

\author{Sergio Guadarrama, Ryan Dahl, David Bieber, Mohammad Norouzi, Jonathon Shlens, Kevin Murphy \\
Google Research \\
\texttt{\{sguada,rld,dbieber,mnorouzi,shlens,kpmurphy\}@google.com} \\
}

\newcommand{\eat}[1]{}
\usepackage{xspace}

\def\eg{\emph{e.g.,}\xspace}

\def\longname{Pixel Recursive Colorization\xspace}
\def\shortname{PixColor\xspace}

\newcommand{\figref}[1]{Figure~\ref{#1}}

\def\ycbcr{YCbCr\xspace} 

\iclrfinalcopy

\begin{document}

\thispagestyle{plain}
\maketitle

\begin{abstract}
We propose a novel approach
to automatically produce multiple colorized versions of a grayscale image.
Our method results from the observation that the task of automated colorization
is relatively easy given a low-resolution version of the color image.
We first train a conditional PixelCNN to generate a low resolution color
for a given grayscale image.
Then, given the generated low-resolution color image and the original grayscale image
as inputs, we train a second CNN to generate a high-resolution
colorization of an image. We demonstrate that our approach
produces more diverse and plausible colorizations than existing methods,
as judged by human raters in a "Visual Turing Test".
\end{abstract}

\section{Introduction}

Building a computer system that can automatically convert a black and
white image to a plausible color image is useful for restoring old photographs,
videos \cite{restoring-conservation},
or even assisting cartoon artists \cite{manga, cartoon}.
From a computer vision perspective,
this may appear like a straightforward image-to-image mapping problem,
amenable to a convolutional neural network (CNN).
We denote this by $y = f(x)$,
where $x$ is the input grayscale image, $y$ is the predicted color image,
and $f$ is a CNN.
This approach has been pursued in several recent papers
\cite{deepcolorization2016, IizukaSIGGRAPH2016, larsson2016learning,
zhang2016colorful, Deshpande-ICCV15, dahl2016, pix2pix2016}
which leverages the fact that one may obtain unlimited labeled training pairs
by converting color images to grayscale.

\begin{figure}[!ht]
\begin{center}
\includegraphics[width=\linewidth]{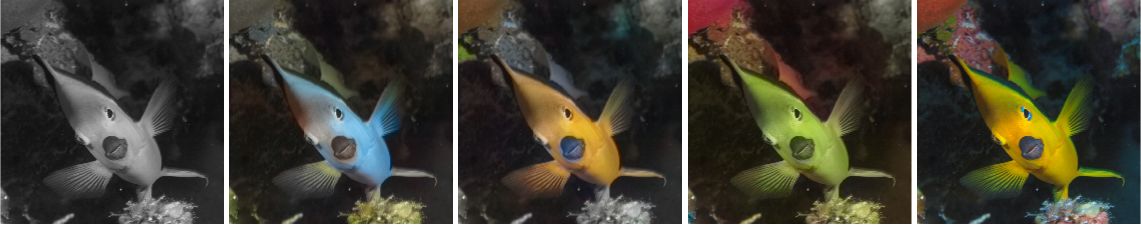}
\vspace{-0.2cm}
   \caption{Grayscale on the left with three colorizations from our model and the original.}
   \label{fig:one-to-many}
\end{center}
\vspace{-0.2cm}
\end{figure}

Removing the chromaticity from an image is a surjective operation, thus restoring
color to an image is a one-to-many operation
(Figure \ref{fig:one-to-many}). We can express this ambiguity as a
conditional probability model
$y \sim p\ ( y\mid x)$ to capture multiple possible outputs,
rather than predicting a single image (see Section \ref{sec:related} for review
of generative models).

In this paper, we propose a new method,
that employs a PixelCNN \cite{pixelcnn} probabilistic model to produce a coherent joint distribution
over color images given a grayscale input.
PixelCNNs have several advantages over other conditional generative models:
(1) they capture dependencies between the pixels to ensure that
colors are selected consistently; (2) the log-likelihood can be computed exactly
and training is stable unlike other generative models.

The main disadvantage of PixelCNNs, however, is that they are slow to sample from,
due to their inherently sequential (autoregressive) structure.
In this paper we leverage the fact that
the chrominance of an image (especially as perceived by humans) is
of much lower spatial frequency than the luminance.
In fact, some image storage formats, such as JPEG, exploit this intuition
and store the color channels at lower resolution than the intensity channel.
This means that it is sufficient for the PixelCNN to predict a low resolution color image,
which may be done quite quickly.
We then train a second CNN-based ``refinement network'',
which combines the predicted low resolution color image
with the high resolution grayscale input
to produce a high resolution color image.

Formally, our approach can be thought of as a conditional latent variable
model of the form $p(y\mid x) = \sum_z  \delta(y = f(x,z)) p(z\mid x)$,
where $x$ is the input grayscale image,
$y$ is the output color image,
$z$ is the latent low-dimensional color image.
The PixelCNN estimates $p(z \mid x)$,
and the refinement CNN estimates $y=f(x,z)$.
At test time, rather than summing over $z$'s, we sample a few $z's$.
During training, we use the ground truth low resolution color image
for $z$, so that we can fit the two conditional models independently.
See Section~\ref{sec:methods} for the details.

\begin{figure}[t]
\begin{center}
\includegraphics[width=\linewidth]{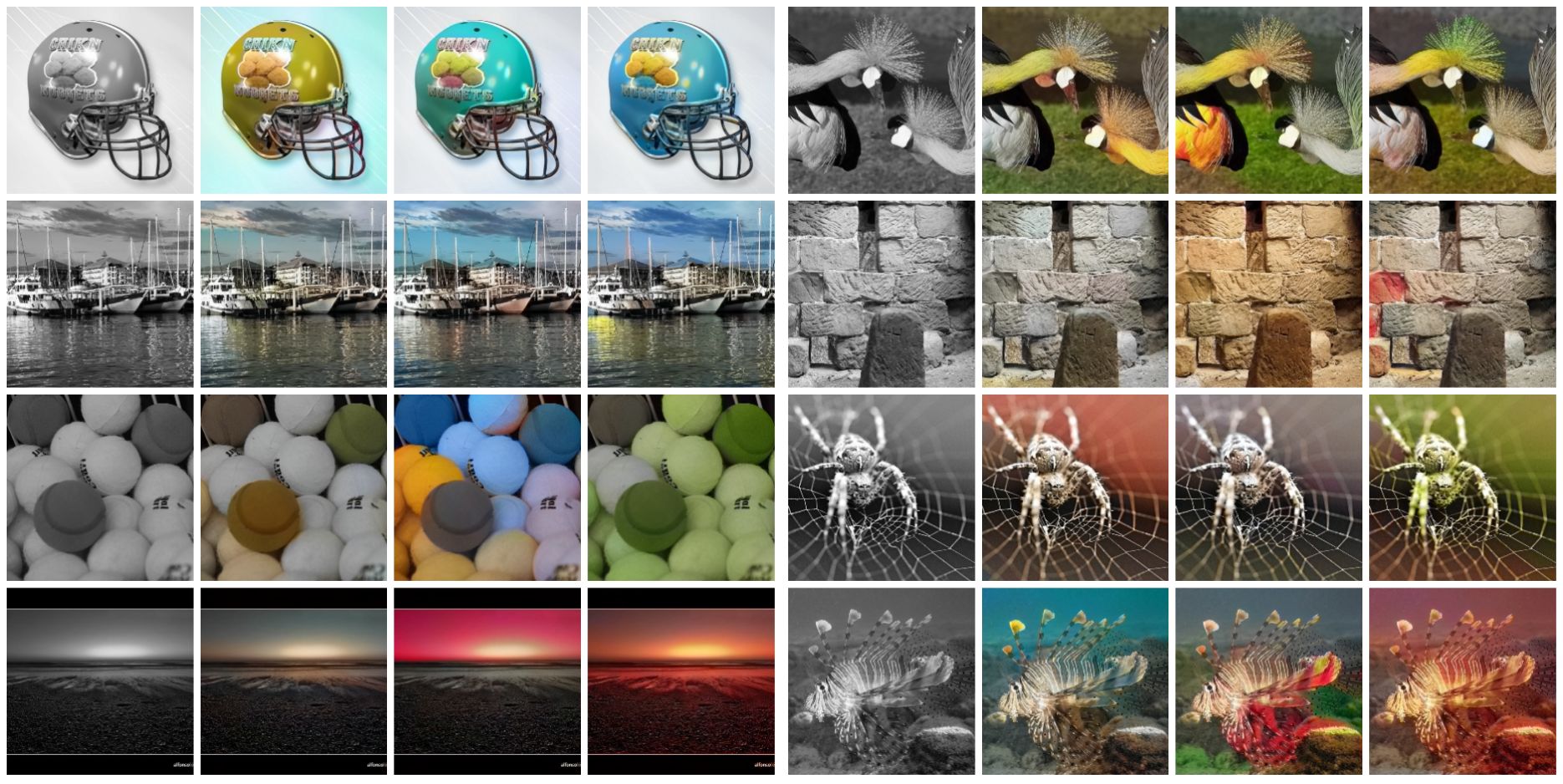}
\end{center}
\vspace{-0.2cm}
   \caption{Diverse colorizations generated by our \shortname method.
   For each group of $4$ images, the first is the grayscale input, and
   the rest are $3$ samples from the model.}
   \label{fig:diverse}
\end{figure}

Our proposed method, called \longname (\shortname), produces diverse,
high quality colorizations. \figref{fig:diverse} depicts some examples
with high diversity.  In Section~\ref{sec:eval}, we describe how we
quantitatively evaluate the performance of colorization using human raters.  We report
our results in Section~\ref{sec:results}, where we show that \shortname
significantly outperforms existing methods. Section~\ref{sec:concl} concludes
the paper and discusses some future directions.

\section{Related work}
\label{sec:related}

Early approaches to colorization relied on some amount of human effort,
either to identify a relevant source color image from which
the colors could be transferred
\cite{welsh2002, chia-semantic-segmentation-2011, gupta2012, irony2005, morimoto-2009,
tai2005, Reinhard-2001,Pitie-2007,charpiat-multimodal},
or to get a rough coloring from a human annotator
to serve as a set of "hints"
\cite{levin04, huang2005, luan2007, manga, yatziv2006, Zhang17,Frans2017}.
More recently, there has been a surge of interest in developing fully automated solutions,
which do not require human interaction (see Table~\ref{tab:comparison}).

Most recent methods train a CNN to map a gray input image to a
single color image~\cite{deepcolorization2016,
  IizukaSIGGRAPH2016, larsson2016learning, zhang2016colorful,
  Deshpande-ICCV15, dahl2016}. When such models are trained with L2 or
L1 loss, the colorization results often look somewhat "washed out",
since the model is encouraged to predict the average color.
Some recent papers
(\eg~\cite{larsson2016learning, zhang2016colorful})
discretize the color space, and use a per-pixel cross-entropy loss on the softmax outputs of a CNN,
resulting in more colorful pictures,
especially if rare colors are upweighted during training (\eg~\cite{zhang2016colorful}).
However, since the model predicts each pixel independently,
the one-to-many nature of the task is not captured properly,
\eg~all of the pixels in a region cannot be constrained to have the same color.

Previous work has proposed several ways to ensure that multiple colorizations
generated by a model are globally coherent.  One approach is to use a
conditional random field (CRF) \cite{charpiat-multimodal},
although inference in such models can be slow.
A second
approach is to use a CNN with multiple output ``heads'',
corresponding to different colorizations of an image.  One can
additionally train a ``gating'' network to select the best
head for a given image.  This mixture of experts (MOE) approach was
used in \cite{Baig2017} mainly for image compression, rather than
colorization {\em per se}.

A third approach is to use a (conditional) variational autoencoder
(VAE) \cite{vae} to capture dependencies amongst outputs
via a low dimensional latent space.
To capture
the dependence on the input image, \cite{diversecolorization2016}
proposes to use a mixture density network (MDN) to learn a mapping
from a gray input image to a distribution over the latent codes, which
is then converted to a color image using the VAE's decoder.
Unfortunately, this method often produces sepia toned results 
(Table~\ref{tab:SxS}).


\begin{table}
\begin{center}
\small
\begin{tabular}{llllllll}
  Name/Ref. & Model & Color  & Loss & Multi &  Dataset  \\ \hline
  AICMP \cite{charpiat-multimodal} & CRF & Lab & N/A & N & N/A \\
  LTBC \cite{IizukaSIGGRAPH2016} & CNN & Lab & L2 + class CE & N  & MIT places \\
  LRAC \cite{larsson2016learning} & CNN & Lab   & CE & N   & ImageNet \\
  CIC \cite{zhang2016colorful} & CNN & Lab & CE & N &  ImageNet  \\
  MOE \cite{Baig2017} & MOE & \ycbcr & L2 & Y &  ImageNet \\
  VAE \cite{diversecolorization2016} & MDN + VAE & Lab & Mahal. & Y  & ImageNet \\
  Pix2Pix \cite{pix2pix2016} & GAN & Lab & Adv. & N &  ImageNet \\
  GAN \cite{Cao2017} & GAN & YUV & Adv. & N & LSUN  \\
  \shortname (this paper) & PixelCNN + CNN & \ycbcr & CE + L1 & Y & ImageNet
\end{tabular}
\end{center}
\caption{
  \footnotesize
  Summary of related methods. Columns comprise
  name of method;
  reference;
  model type (MOE = mixture of experts, VAE = variational autoencoder,
  MDN = mixture density network,
  GAN = generative adversarial network);
  color space;
  loss (CE = cross entropy, Mahal = Mahalanobis distance, Adv = adversarial);
  multiple diverse outputs or not; and the dataset used to train the model.
  The CRF method of \cite{charpiat-multimodal} requires that the user specify
  one or more training images that are similar to the input gray image.
  Although the CRF is is
  capable of generating
  multiple solutions, \cite{charpiat-multimodal} uses graph-cuts to produce a single MAP estimate.
  Similarly, although the GAN method of \cite{pix2pix2016} is capable of producing multiple solutions,
  they report that their GAN ignores the noise, and always predicts the same answer for each input.
  This problem is fixed in \cite{Cao2017} by introducing noise at multiple levels of the generator.
}
\label{tab:comparison}
\end{table}

A fourth approach is to use a (conditional) generative adversarial
network (GAN) \cite{gan} to train a generative model jointly with
a discriminative model. The goal of the discriminative model is detect
synthesized images, while the goal of the generative model is a fool
the discriminator.
This approach results sharp images, but \cite{pix2pix2016} reports that a
GAN-based colorization results underperform previous CNN approaches
\cite{zhang2016colorful}. One of their failure modes
``mode collapse'' problem, whereby the resulting model correctly predicts
one mode of a distribution but fails the full diversity of the data \cite{unrolled-gans}.
More recently, \cite{Cao2017} have applied a slightly
different GAN to colorization.  Although the authors claim to avoid the mode collapse
problem, it is hard to compare against previous results because the authors
only employ the LSUN-bedrooms dataset for evaluation. Most papers (including ours)
employ the ``ctest10k'' split of the ImageNet
validation dataset from~\cite{larsson2016learning} (see
Section~\ref{sec:eval} for more details).

We propose a novel approach that uses a PixelCNN~\cite{pixelcnn} to
produce multiple low resolution color images, which are then
deterministically converted to high resolution color images using a
CNN refinement network. By using multiple low resolution color "hints" to
the CNN, we capture the one-to-many nature of the task and prevent
the CNN from producing sepia toned outputs.

Very recently, in a concurrent submission,
\cite{Royer2017} proposed an approach which is similar to ours.
However, instead of passing the output of a pixelCNN into  a refinement CNN,
they do the opposite, and pass the output a CNN into a pixelCNN.
The visual quality and diversity of their results look good, but,
unlike us, they do not
perform any human evaluation, so we do not have a quantitative
comparison.
The primary disadvantage of their approach is that it is slow for a pixelCNN to generate
high resolution images; indeed, their method only generates
$32 \times 32$ color images, which are then deterministically upscaled to $128 \times 128$.
By contrast,
our CNN refinement network learns to upscale from $28 \times 28$ to the
same size as the input, which works much better than deterministic upscaling,
as we will show.
We mostly focus on generating $256 \times 256$ images, to be comparable to prior work,
but we also show some non-square examples, which is important in practice,
since many grayscale photos of interest are in portrait or landscape mode.


\section{\longname (\shortname)}
\label{sec:methods}

\begin{figure}
\begin{center}
   \includegraphics[width=\linewidth]{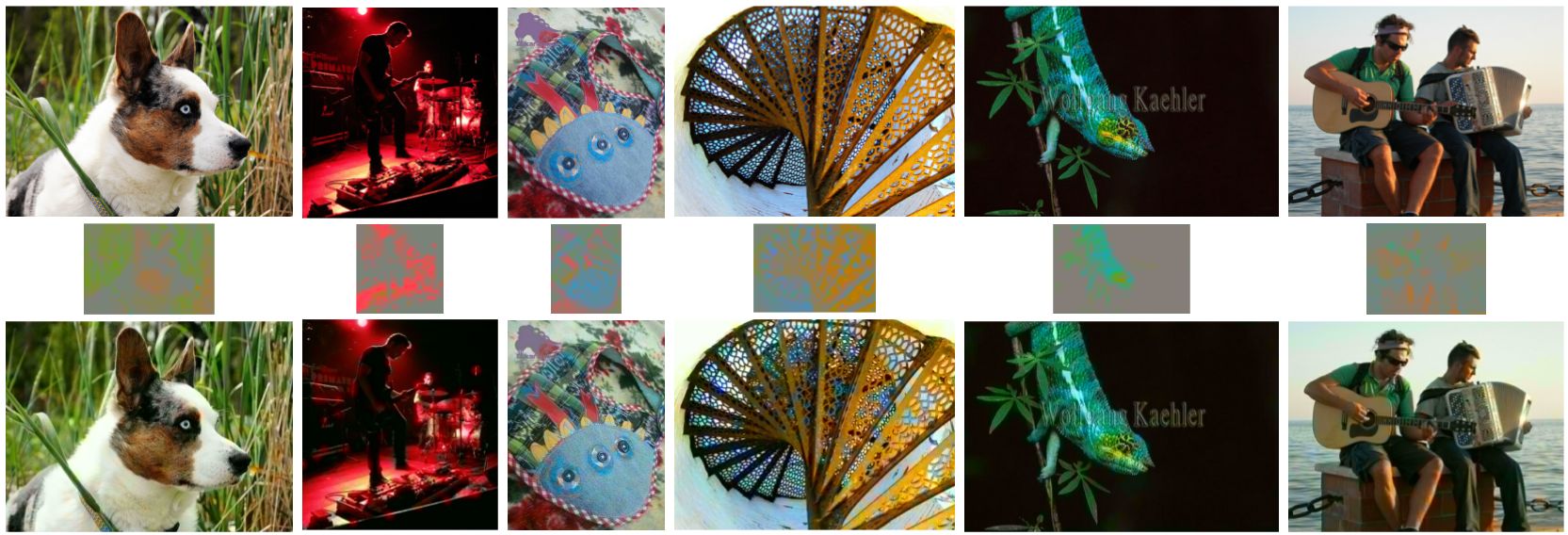}
\end{center}
   \caption{  \footnotesize
   All you need is a few bits of color.
   The top row is the original color image. The middle row is the true
   chroma image downsampled to have smallest size 28 pixels. The bottom row is the result of
   bilinear upsampling the middle row, and combining with the original grayscale image.
   }
\label{fig:upsample}
\end{figure}

The key intuition behind our approach is that it suffices
to predict a plausible {\em low resolution} color image,
since color is much lower spatial frequency than intensity.
To illustrate this point, suppose we take
the \emph{ground truth} chrominance of an image, downsample it to
$28\!\times\!28$, upsample it back to the original size, and then
combine it with the original luminance.
\figref{fig:upsample} shows some examples of this process.
It is clear that the resulting
colorized images look very close to the original color images.

\begin{figure}
\begin{center}
   \includegraphics[width=0.9\linewidth]{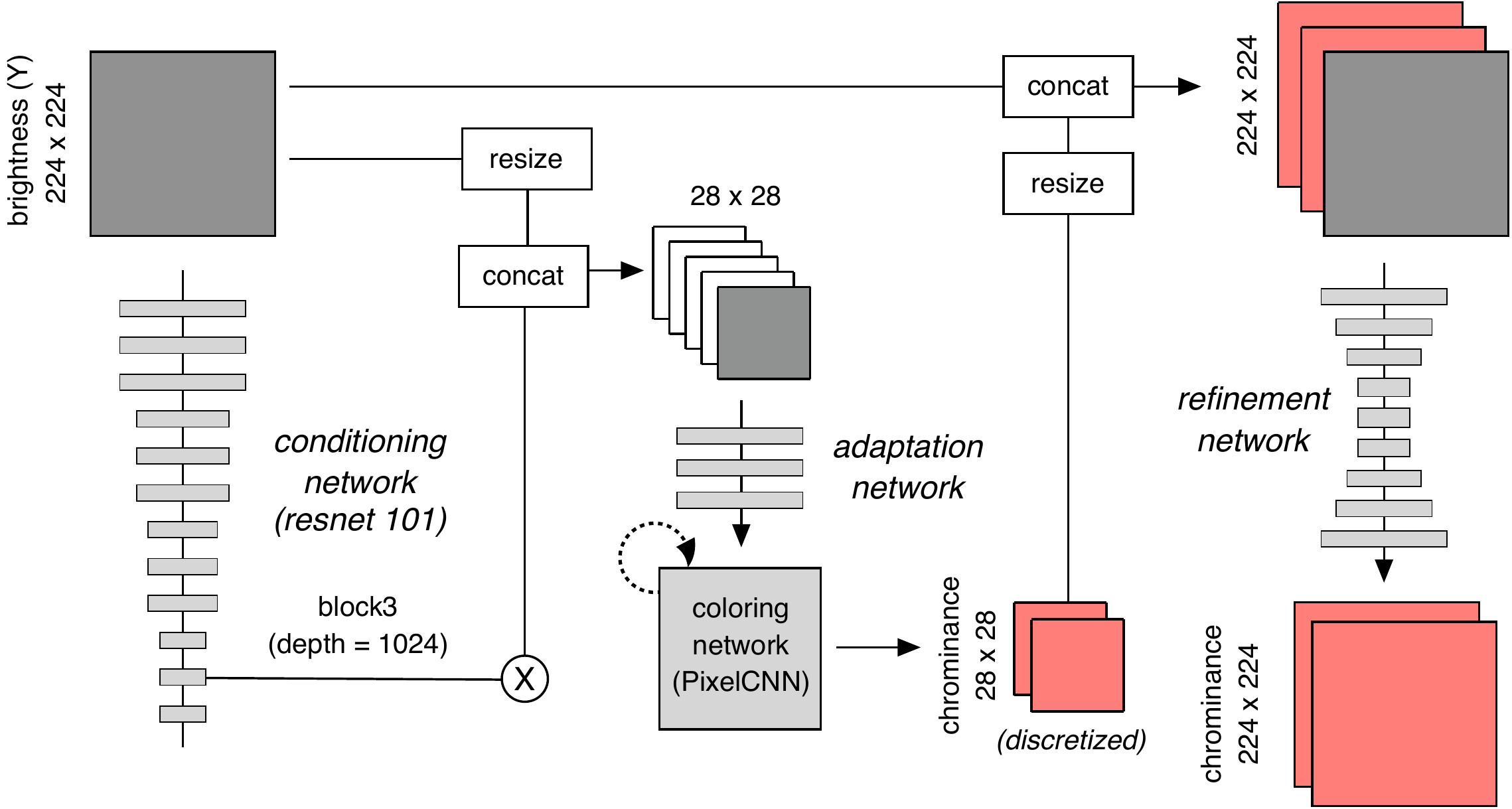}
\end{center}
   \caption{
  \footnotesize
   Diagram of \longname (\shortname) method.
   We first pre-train the {\it conditioning network} on COCO image segmentation
   following \cite{chen2016deeplab}.
   Then, the conditioning network and the {\it adaptation network} convert the
   brightness channel $Y$ into a set of features providing the necessary
   conditioning signal to the PixelCNN.
   The PixelCNN is optimized jointly with the conditioning and adaptation
   networks to predict a low spatial resolution
   version of the color image in a discretized space.
   The low spatial resolution image is subsequently
   supplied to a {\it refinement network}, which is trained to produce a full
   resolution colorization.}
\label{fig:overview-diagram}
\end{figure}

In the sections below, we describe how we train
a model to predict multiple plausible low resolution color images,
and then how we train a second model to combine these predictions
with the original grayscale input to produce a high resolution
color output.
See \figref{fig:overview-diagram} for an overview the
approach.

\subsection{PixelCNN for low-resolution colorization}
\label{sec:pixelCNN}

Inspired by the success of autoregressive models for
unconditional image generation \cite{pixelrnn,pixelcnn}
and super resolution \cite{Dahl2017}, we
use a conditional PixelCNN \cite{pixelcnn} to produce multiple
low resolution color images.
That is, we
turn colorization into
a sequential decision making task, where pixels are colored
sequentially, and the color of each pixel is conditioned on the input
image and previously colored pixels.

Although sampling from a PixelCNN is in general quite slow (since it is
inherently sequential), we only need to generate a low-resolution
image (28x28), which is reasonably fast.
In addition, there are various additional speedup tricks we can use
(see e.g., \cite{Ramachandran2017,Kolesnikov2016})
if necessary.

Our architecture is based on \cite{Dahl2017}
who used PixelCNNs
to perform super resolution (another one-to-many problem).
We use the \ycbcr colorspace, because it is linear, simple and widely used
(e.g., by JPEG).
We discretize the Cb and Cr channels separately into 32 bins.
Thus the model has the following form:
\[
p(y|x) =
  \prod_i p(y(i,r) \mid y(1:i-1,:),\, x) \,
  p(y(i,b) \mid y(i,r),\, y(1:i-1, :),\, x)
\]
where $y(i,r)$ is the Cr value for pixel $i$,
and $y(i,b)$ is the Cb value.
We performed some preliminary experiments using Logistic mixture models
to represent the output values as suggested by the PixelCNN++ of~\cite{Salimans2017},
as opposed to using multinomials over discrete values~\cite{pixelcnn}.
However, we did not see a meaningful improvement, so for simplicity,
we stick to a multinomial prediction model.

We train this model using maximum likelihood, with a cross-entropy loss per pixel.
Because of the sequential nature of the model, each prediction is conditioned on previous pixels.
During training, we "clamp" all the previous pixels to the ground truth values
(an approach known as "teacher forcing" \cite{Williams1989}), and just train 
the network to predict a single pixel at a time.
This can be done efficiently in parallel across pixels.

\subsection{Feedforward CNN for high-resolution refinement}
\label{sec:refinement}

\begin{figure}
  \centering
  \begin{tabular}{m{0.6\linewidth}m{0.4\linewidth}}
\def \w { 1.9cm }
\def \s { .12cm }
\begin{tabular}{@{\hspace{\s}}c@{\hspace{\s}}c@{\hspace{\s}}c@{\hspace{\s}}c@{\hspace{\s}}c}
 Input &  Sample & Refined & Output \\
{\includegraphics[width=\w]{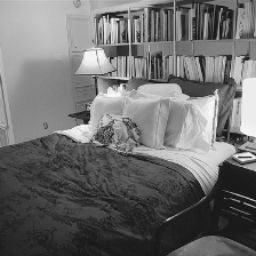}}
 & {\includegraphics[width=\w]{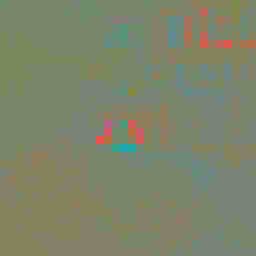}}
 & {\includegraphics[width=\w]{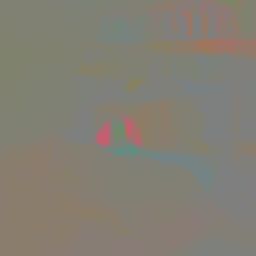}}
 & {\includegraphics[width=\w]{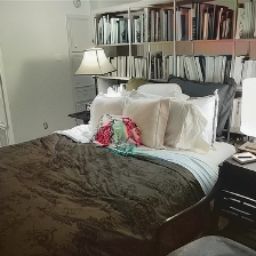}}
 \\
{\includegraphics[width=\w]{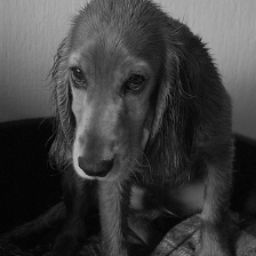}}
 & {\includegraphics[width=\w]{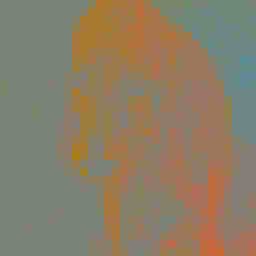}}
 & {\includegraphics[width=\w]{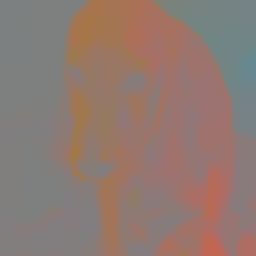}}
 & {\includegraphics[width=\w]{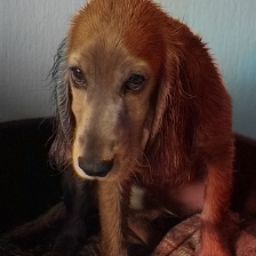}}
 \\
{\includegraphics[width=\w]{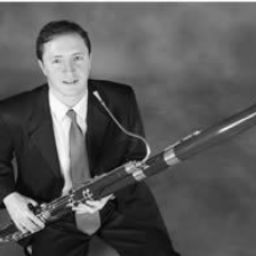}}
 & {\includegraphics[width=\w]{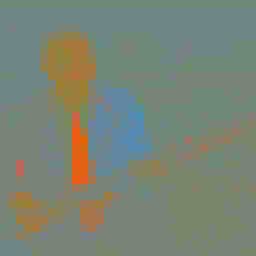}}
 & {\includegraphics[width=\w]{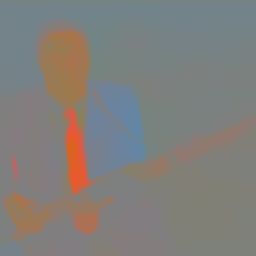}}
 & {\includegraphics[width=\w]{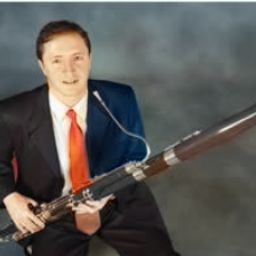}}
 \\
\end{tabular}

    &
    \begin{tabular}{
    @{\hspace{.1cm}}c|
    @{\hspace{.1cm}}c|
    @{\hspace{.1cm}}c
  }
  \multicolumn{3}{c}{Ablation Study} \\
  [.1cm]
   & Sample & GT 28x28 \\
  \hline
  Unrefined & 19.9\% & 29.6\% \\
  Refined & 33.9\% & 43.6\%
\end{tabular}

  \end{tabular}
  \caption{\small
  Left: The intermediate stages of \shortname.
  The column labeled "sample" is an output from PixelCNN, upsampled to the size
  of the image for visualization purposes.
  The column labeled "refined" is the output of the refinement network,
  before being combined with the grayscale input.
  Right:
  Results of a human evaluation using the "Visual Turing Test" metric
  explained in Section~\ref{sec:eval}.
  We compare four systems: ground truth (GT) chroma image vs generated
  sample, passed directly into bilinear upsampling (unrefined)
  or passed into the refinement network.
  }
  \label{fig:refinement}
\end{figure}

A simple way to use the low resolution output of the colorization network
is to upsample it (e.g., using bilinear or nearest neighbor interpolation),
and then to concatenate the result with the original luminance channel.
This can work quite well given groundtruth color, as we showed in Figure~\ref{fig:upsample}.
However, it is possible to do better by learning how to combine the
predicted low resolution color image with the original high resolution grayscale image.

For this, we use an image-to-image CNN which we call the refinement network.
It is similar in architecture to the network used in \cite{IizukaSIGGRAPH2016}
but with more layers in the decoding part. In addition,
we use bilinear interpolation for
upsampling instead of learned upsampling.

The refinement network is trained on a 28x28 downsampling of the ground truth chroma
images. The reason we do not train it end-to-end with the PixelCNN is the following:
the PixelCNN can generate multiple samples, all of which might be quite far
from the true chroma image; if we forced the refinement network to map these
to the true RGB image, it might learn to ignore these "irrelevant" color "hints",
and just use the input grayscale image. By contrast, when we train using the true
low-resolution chroma images, we force the refinement network to focus its efforts on learning
how to combine these "hints" with the edge boundaries which are encoded in the grayscale image.

We show some qualitative examples of the benefits of the refinement network
on the left of Figure~\ref{fig:refinement}.
At first glance, the benefits seem small, but if you zoom in you will notice
that the refinement network's outputs are much more plausible, since they
better adhere to segment boundaries, etc.
The results of a quantitative human evaluation
of the refinement network,
using the "Visual Turing Test" metric
  explained in Section~\ref{sec:eval},
are shown in the table on the right of
Figure~\ref{fig:refinement}.
 The increase from the
  Sample-Unrefined score (19.9\%)
  to the Sample-Refined score (33.9\%) shows the value added by the refinement
  network.
  The GT-Refined score (43.6\%) shows the upper limit of our method could
  achieve with our refinement network (the maximum expected score for VTT
  is 50\%).

\section{Evaluation methodology}
\label{sec:eval}

Since the mapping from gray to color is one-to-many,
we cannot evaluate performance by
comparing the predicted color image to the
"ground truth" color image in terms of mean squared error or even other
perceptual similarity metrics such as SSIM \cite{Wang2003ssim}.
Instead, we follow the approach of \cite{zhang2016colorful}
and conduct a "Visual Turing Test" (VTT) using a crowd sourced human raters.
In this test,
we present two different color versions of an image, one the ground truth and
one corresponding to the predicted colors generated by some method.
We then ask the rater to pick the image which has the "true colors".
A method that always produces the ground truth colorization would score 50\%
by this metric.

To be comparable with \cite{zhang2016colorful},
we show the two images sequentially for 1 second each.
(We randomize which image is shown first.)
Following standard practice, we train on the
1.2M training images from the ILSVRC-CLS dataset
\cite{ILSVRC15}, and use 500 images from the
"ctest10k" split
of the 50k ILSVRC-CLS validation dataset
proposed in \cite{larsson2016learning}.
Each image is shown to 5 different raters.
We then compute the fraction of times the generated image is preferred
to ground truth; we will call this the "VTT score" for short.

\section{Results}
\label{sec:results}

We assess the effectiveness of our technique by comparing against
several recent colorization methods, both qualitatively and
quantitatively. Table~\ref{tab:SxS} shows a qualitative comparison
of various recent methods applied to a few randomly chosen test
images.
Based on these examples, it seems that the best methods include our
method (\shortname), and several recent CNN-based methods, namely LTBC
\cite{IizukaSIGGRAPH2016}, LRAC \cite{larsson2016learning}, and CIC
\cite{zhang2016colorful}. Therefore, we conduct a more costly "Visual
Turing Test" (VTT) on these four systems, as explained in
Section~\ref{sec:eval}.


Figure~\ref{fig:vtt} summarizes the VTT scores.
We see that our method significantly outperforms the previous state of the
art methods, with an average VTT score of 33.9\%.

\begin{figure}
  \centering
\begin{tabular}{c||c|c|c|c|c|c|c}
\multirow{2}{*}{Method} & \multirow{2}{*}{LTBC} & \multirow{2}{*}{CIC} & \multirow{2}{*}{LRAC} & \shortname & \shortname & \shortname & \shortname \\
 & & & & (Seed 1) & (Seed 2) & (Seed 3) & (Oracle) \\
\hline
\multirow{2}{*}{\raisebox{-0.5mm}{VTT (\%)}} & 25.8 & 29.2 & 30.9 & 33.3 & {35.4} & 33.2 & \bf{38.3} \\
 & $\pm 0.97$ & $\pm 0.98$ & $\pm 1.02$ & $\pm 1.04$ & $\pm 1.01$ & $\pm 1.03$ & $\pm 0.98$ \\
\end{tabular}

   \caption{
   Results of the Visual Turing Test (VTT) study on the ImageNet test set.
   We report the fraction of times raters picked the generated color image
   over the ground truth with error ranges produced by
   bootstrapping the mean.
   Our study includes $500$ test images and $5$ raters per image.
   The column labeled "Oracle" is the score of the single best sample
   per image chosen by human raters. 
  }
  \label{fig:vtt}
\end{figure}

\begin{figure}
  \centering
  \begin{tabular}{ m{0.6\linewidth} m{0.4\linewidth}}
    \includegraphics[width=\linewidth]{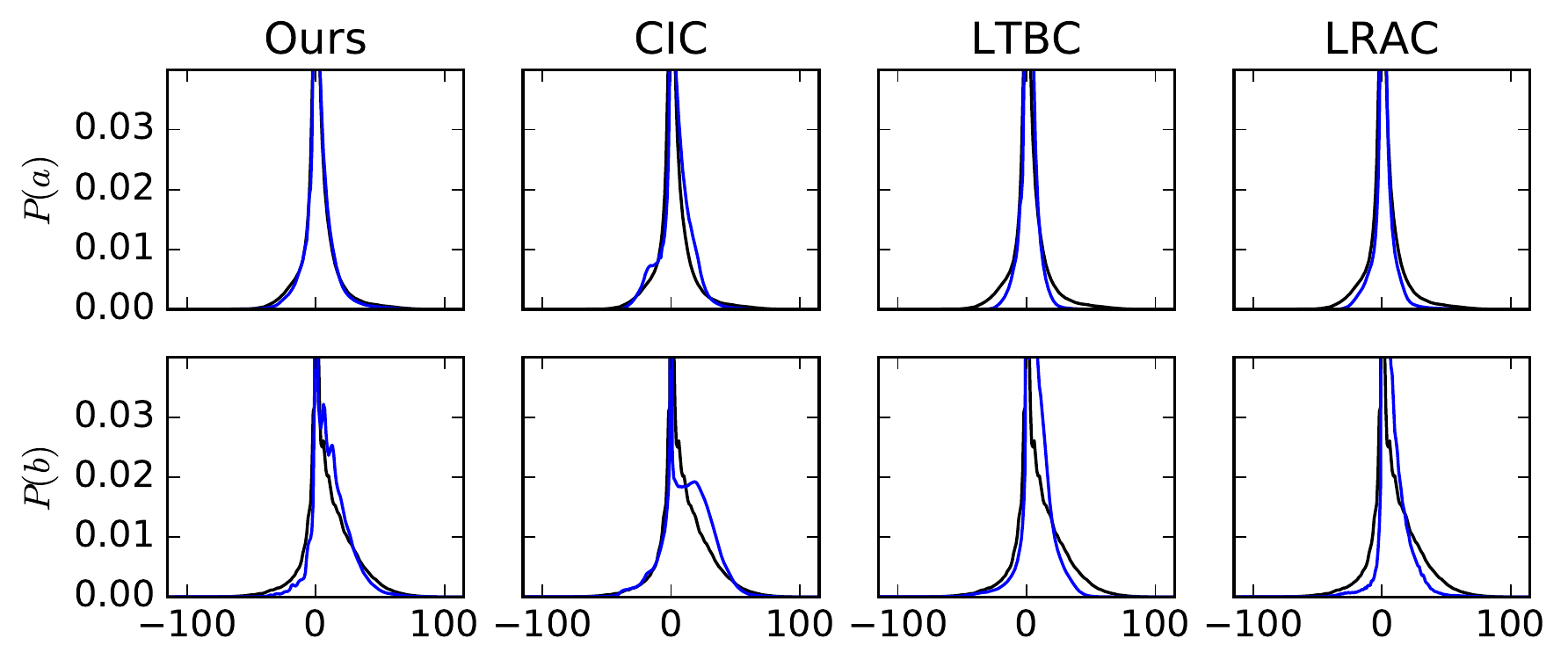}
    &
    \begin{tabular}{r|c|c}
      \multicolumn{3}{c}{Histogram Intersection} \\[.1cm]
       & a & b \\
        \hline
      \shortname & \bf{0.93} & \bf{0.93} \\
             CIC & 0.85 & 0.85 \\
            LTBC & 0.82 & 0.82 \\
            LRAC & 0.78 & 0.78 \\
      \end{tabular}
  \end{tabular}
  \caption{Marginal statistics of the color channels in Lab color space.
   Left: each method's histogram is shown in blue against ImageNet's test set
   histogram in black. Right: Histogram intersection on the color channels.}
  \label{fig:marginals}
\end{figure}

One reason we think our results are better is that the colors they produce
are more "natural", and are placed in the "right" places.
To assess the first issue,
Figure~\ref{fig:marginals} plots the marginal statistics of the
$a$ and $b$ channels (of CIELab) derived from the images generated from each image.
We see that our model matches the empirical distribution (derived
from the true color images) more closely than the other methods,
without needing to do any explicit reweighting of color bins,
as was done in previous work \cite{zhang2016colorful}.

\subsection{Sample diversity}
\label{sec:diversity}

Our model can produce multiple samples for each input,
so for we run it 3 times, with 3 different seeds, and evaluate the outputs
of each run independently.
From Figure~\ref{fig:vtt}, we see that all of the samples
are fairly good, but are they different from each other?
That is, are the samples diverse?

\begin{figure}
  \centering
\def \w { 2.68cm }
\def \s { .15cm }
\centering
\begin{tabular*}{\textwidth}{
  @{\hspace{\s}}c
  @{\hspace{\s}}c
  @{\hspace{\s}}c
  @{\hspace{\s}}c
  @{\hspace{\s}}c
}

  SSIM = 0.80 & 
  SSIM = 0.85 & 
  SSIM = 0.90 & 
  SSIM = 0.95 & 
  SSIM = 0.99 \\

\includegraphics[width=\w]{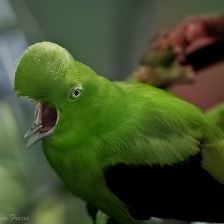}  &  
\includegraphics[width=\w]{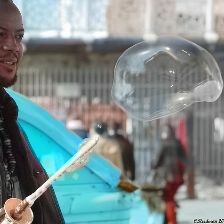}  &  
\includegraphics[width=\w]{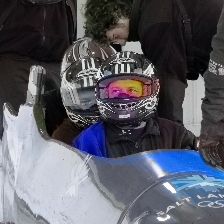}  &  
\includegraphics[width=\w]{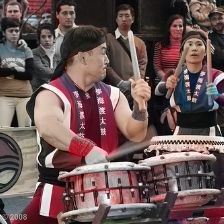}  &  
\includegraphics[width=\w]{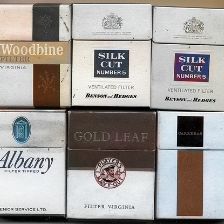}  \\

\includegraphics[width=\w]{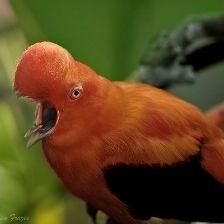}  &  
\includegraphics[width=\w]{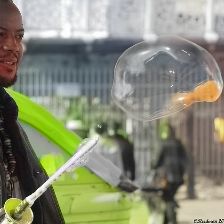}  &  
\includegraphics[width=\w]{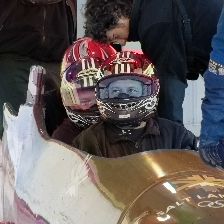}  &  
\includegraphics[width=\w]{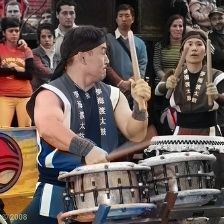}  &  
\includegraphics[width=\w]{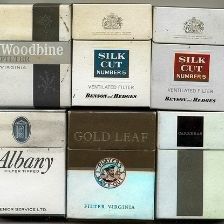}  \\  

  \multicolumn{5}{c}{
    \includegraphics[width=\linewidth]{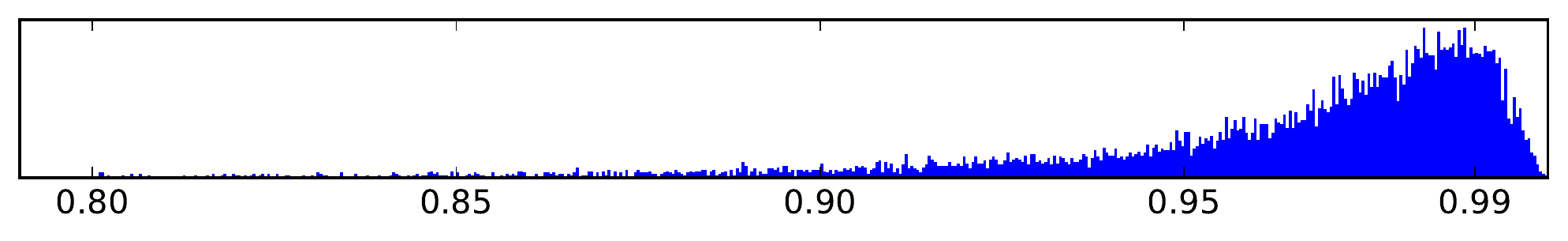}
  } 
  
\end{tabular*}

  \caption{
To demonstrate that our model produces diverse samples, we compare two outputs
from the same input with multiscale SSIM. A histogram of the SSIM
distances from the ImageNet test set is shown above. Representative pairs are
shown at at various SSIM distances. An SSIM value of 1.0 means the images are
identical.
  }
  \label{fig:diversity}
\end{figure}

Figure~\ref{fig:diverse} suggests that our method can generate diverse samples.
To quantitatively assess how different these samples are from each other,
we compute the multiscale SSIM \cite{Wang2003ssim} measure between
pairs of samples. The results are shown in Figure~\ref{fig:diversity}.
We see that most pairs have an SSIM score in the 0.95-0.99 range,
meaning that they are very similar, but differ in a few places,
corresponding to subtle details, such as the color of a person's shirt.
The pairs which have the lowest SSIM score are the ones where large
objects are given different colors (see the pair of birds on the left hand
side).

In an ideal world, we could automatically select the single best sample,
and just show that to the user. To get a sense of how well this could perform,
we decided to use
humans to perform the task of picking the best sample.
More precisely, for each of the 3 samples for a given image,
we picked the one that the most raters liked.
We then computed the VTT score for these single samples using a different
set of raters.
The VTT score jumps to $38.3\%$. This suggests that an algorithmic way to pick
a good sample from the set could yield significantly better results.

We did some preliminary experiments
where we used the likelihood score
(according to the PixelCNN model) to pick the best sample,
but this did not yield good correlation with human judgement.
It may be possible to train a separate ranking model, but we leave
that to future work.

\vspace*{-.1cm}
\section{Conclusion}
\vspace*{-.1cm}
\label{sec:concl}

\begin{figure}
  \centering
  \newcommand\raisescore[1]{\raisebox{9mm}{#1}}

\def \w { .13 }

\begin{tabular}{@{\hspace{.1cm}}c@{\hspace{.1cm}}c@{\hspace{.1cm}}c@{\hspace{.1cm}}c@{\hspace{.1cm}}c@{\hspace{.1cm}}c@{\hspace{.1cm}}c}
VTT Score & \shortname & G. Truth & \shortname & G. Truth & \shortname & G. Truth \\
\raisescore{0\%} & {\includegraphics[width=.13\linewidth]{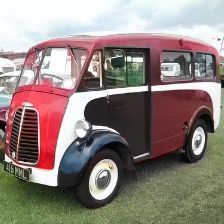}}
 & {\includegraphics[width=\w\linewidth]{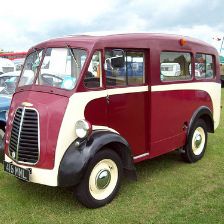}}
 & {\includegraphics[width=\w\linewidth]{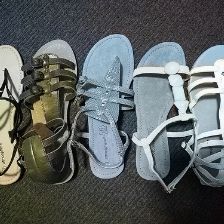}}
 & {\includegraphics[width=\w\linewidth]{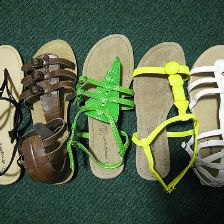}}
 & {\includegraphics[width=\w\linewidth]{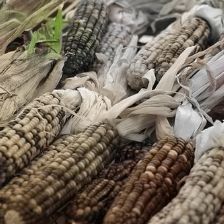}}
 & {\includegraphics[width=\w\linewidth]{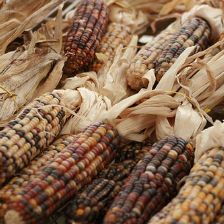}}
 \\
\raisescore{20\%} & {\includegraphics[width=\w\linewidth]{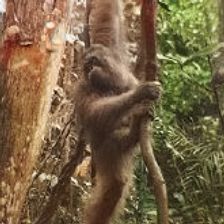}}
 & {\includegraphics[width=\w\linewidth]{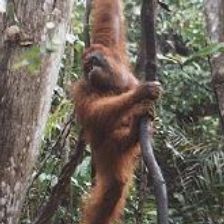}}
 & {\includegraphics[width=\w\linewidth]{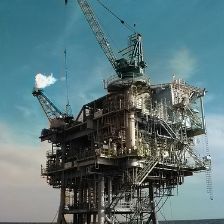}}
 & {\includegraphics[width=\w\linewidth]{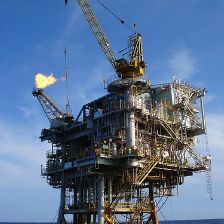}}
 & {\includegraphics[width=\w\linewidth]{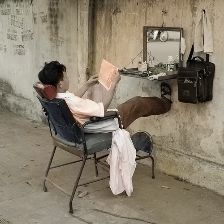}}
 & {\includegraphics[width=\w\linewidth]{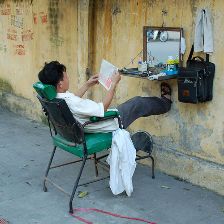}}
 \\
\raisescore{40\%} & {\includegraphics[width=\w\linewidth]{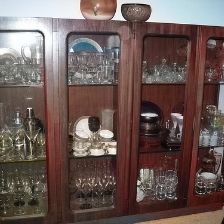}}
 & {\includegraphics[width=\w\linewidth]{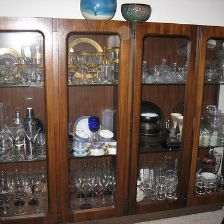}}
 & {\includegraphics[width=\w\linewidth]{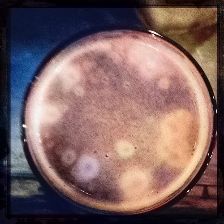}}
 & {\includegraphics[width=\w\linewidth]{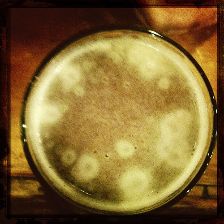}}
 & {\includegraphics[width=\w\linewidth]{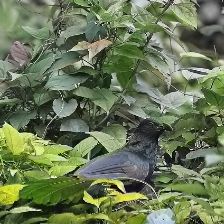}}
 & {\includegraphics[width=\w\linewidth]{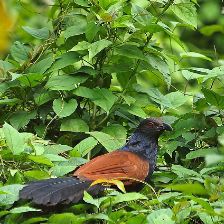}}
 \\
\raisescore{60\%} & {\includegraphics[width=\w\linewidth]{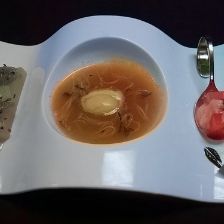}}
 & {\includegraphics[width=\w\linewidth]{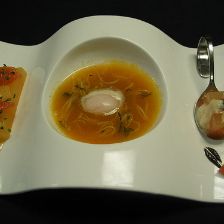}}
 & {\includegraphics[width=\w\linewidth]{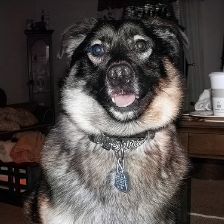}}
 & {\includegraphics[width=\w\linewidth]{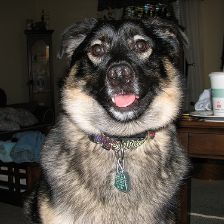}}
 & {\includegraphics[width=\w\linewidth]{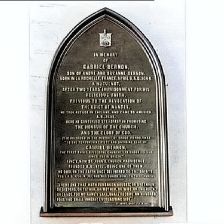}}
 & {\includegraphics[width=\w\linewidth]{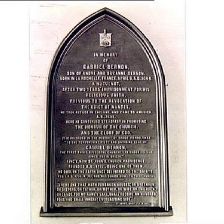}}
 \\
\raisescore{80\%} & {\includegraphics[width=\w\linewidth]{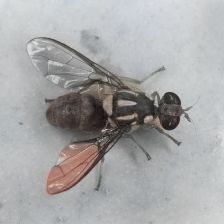}}
 & {\includegraphics[width=\w\linewidth]{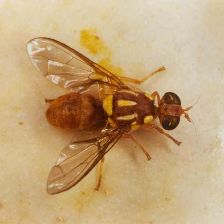}}
 & {\includegraphics[width=\w\linewidth]{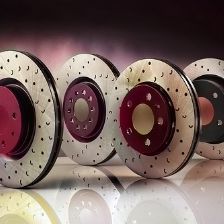}}
 & {\includegraphics[width=\w\linewidth]{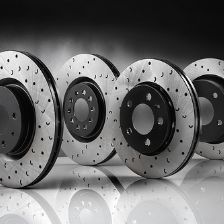}}
 & {\includegraphics[width=\w\linewidth]{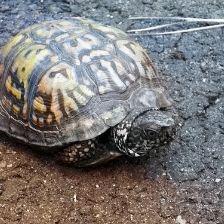}}
 & {\includegraphics[width=\w\linewidth]{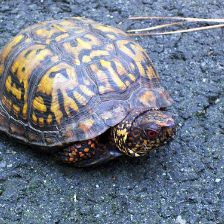}}
 \\
\raisescore{100\%} & {\includegraphics[width=\w\linewidth]{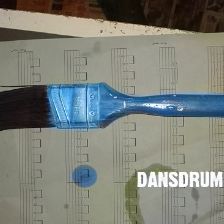}}
 & {\includegraphics[width=\w\linewidth]{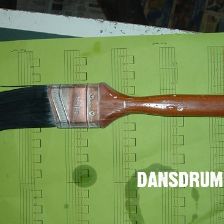}}
 & {\includegraphics[width=\w\linewidth]{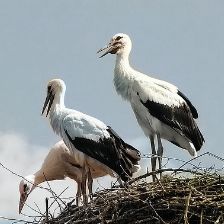}}
 & {\includegraphics[width=\w\linewidth]{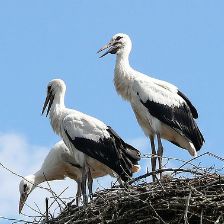}}
 & {\includegraphics[width=\w\linewidth]{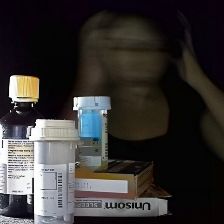}}
 & {\includegraphics[width=\w\linewidth]{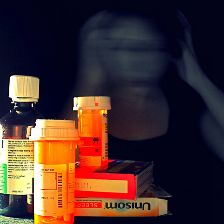}}
 \\
\end{tabular}

  \caption{ Images illustrating each possible VTT score.  }
  \label{fig:bestworse}
\end{figure}

We showed \shortname produces diverse colorizations and found that on average
the outputs of our model perform better than other published methods in a crowd
sourced human evaluation. We avoid the problem of slow inference in PixelCNN
by only sampling low-resolution color channels and use a standard
image-to-image CNN to refine the result. We justified the necessity of the
refinement network with ablation studies and we showed that \shortname
outputs more closely match the marginal color distributions when compared to
other methods.
The model exhibits a variety of failure modes, as illustrated in 
Figure~\ref{fig:bestworse},
which we will address in our future work.

\subsubsection*{Acknowledgments}

We thank Stephen Mussmann and Laurent Dinh for work and discussion on earlier versions of this project; 
Julia Winn, Jingyu Cui and Dhyanesh Narayanan for help with an earlier prototype;
A{\"a}ron van den Oord for advice and guidance employing PixelCNN architectures;
the TensorFlow team for technical and infrastructure assistance.

{
\bibliographystyle{plain}
\bibliography{paper}
}
\raggedbottom

\section*{Appendix}

\begin{table}[H]
  \centering
  \begin{tabular}{@{}rlll@{}} \toprule
Operation               & Kernel       & Strides      & Feature maps  \\
\midrule

\bf PixelCNN conditioning network & \multicolumn{3}{l}{$224 \times 224 \times 3$ input} \\

Conv2D  & $7 \times 7$ & 2 & $64$ \\
$3 \times$ ResNet block   & $3 \times 3$          & 2, 1, 1          & $64/256$ bottleneck  \\
$4 \times$ ResNet block   & $3 \times 3$          & 2, 1, 1, 1          & $128/512$ bottleneck  \\
$23 \times$ ResNet block   & $3 \times 3$          & 1          & $256/1024$ bottleneck  \\
$3 \times$ Conv2D  & $3 \times 3$ & 1 & $64$ \\
Gradient Multipler & \multicolumn{3}{l}{Engaged at 100,000 steps} \\

\midrule
\bf PixelCNN colorization network & \multicolumn{3}{l}{$28 \times 28 \times 2$ input} \\
Masked Conv2D  & $7 \times 7$ & 1 & $64$  \\
10 $\times$ Gated Conv2D Blocks  & $5 \times 5$ & 1 & $64$  \\
Masked Conv2D  & $1 \times 1$ & 1 & $1024$  \\
Masked Conv2D  & $1 \times 1$ & 1 & $2 * 32$  \\

\midrule
\bf Refinement Network & \multicolumn{3}{l}{$224 \times 224 \times 3$ input} \\

Conv2D & $3\times 3$ & 2 & 64 \\
$2\times$ Conv2D & $3\times 3$ & 1, 2 & 128 \\
$2\times$ Conv2D & $3\times 3$ & 1, 2 & 256 \\
$2\times$ Conv2D & $3\times 3$ & 1 & 512 \\
Conv2D & $3\times 3$ & 1 & 256 \\
$4\times$ Conv2D & $3\times 3$ & 2, 1, 2, 1 & 512 \\
Conv2D & $5\times 5$ & 1 & 1024 \\
Conv2D & $3\times 3$ & 1 & 512 \\
$2 \times$ Conv2D & $3\times 3$ & 1 & 128 \\
Bilinear Upsample  \\
$2 \times$ Conv2D & $3\times 3$ & 1 & 64 \\
Bilinear Upsample  \\
$2 \times$ Conv2D & $3\times 3$ & 1 & 32 \\
Conv2D & $1 \times 1$ & 1 & 2 \\

\hline
\hline
Optimizer     & \multicolumn{3}{l}{Adam (beta1=0.9, momentum=0.9)}  \\
Batch size    & \multicolumn{3}{l}{$64 = 8 \times 8$ (8 GPUs, synchronous updates)} \\
Iterations    & \multicolumn{3}{l}{360,000} \\
Learning Rate & \multicolumn{3}{l}{0.0003} \\
Colorness Threshold & \multicolumn{3}{l}{0.05} \\
Weight, bias initialization  & \multicolumn{3}{l}{Truncated normal (stddev=0.1), Constant($0$)} \\ \bottomrule
\end{tabular}

  \caption{Hyperparameters used.
  The PixelCNN and Refinement networks were trained independently.  }
\end{table}

\begin{figure}
  \centering
  \includegraphics[width=.6\paperwidth]{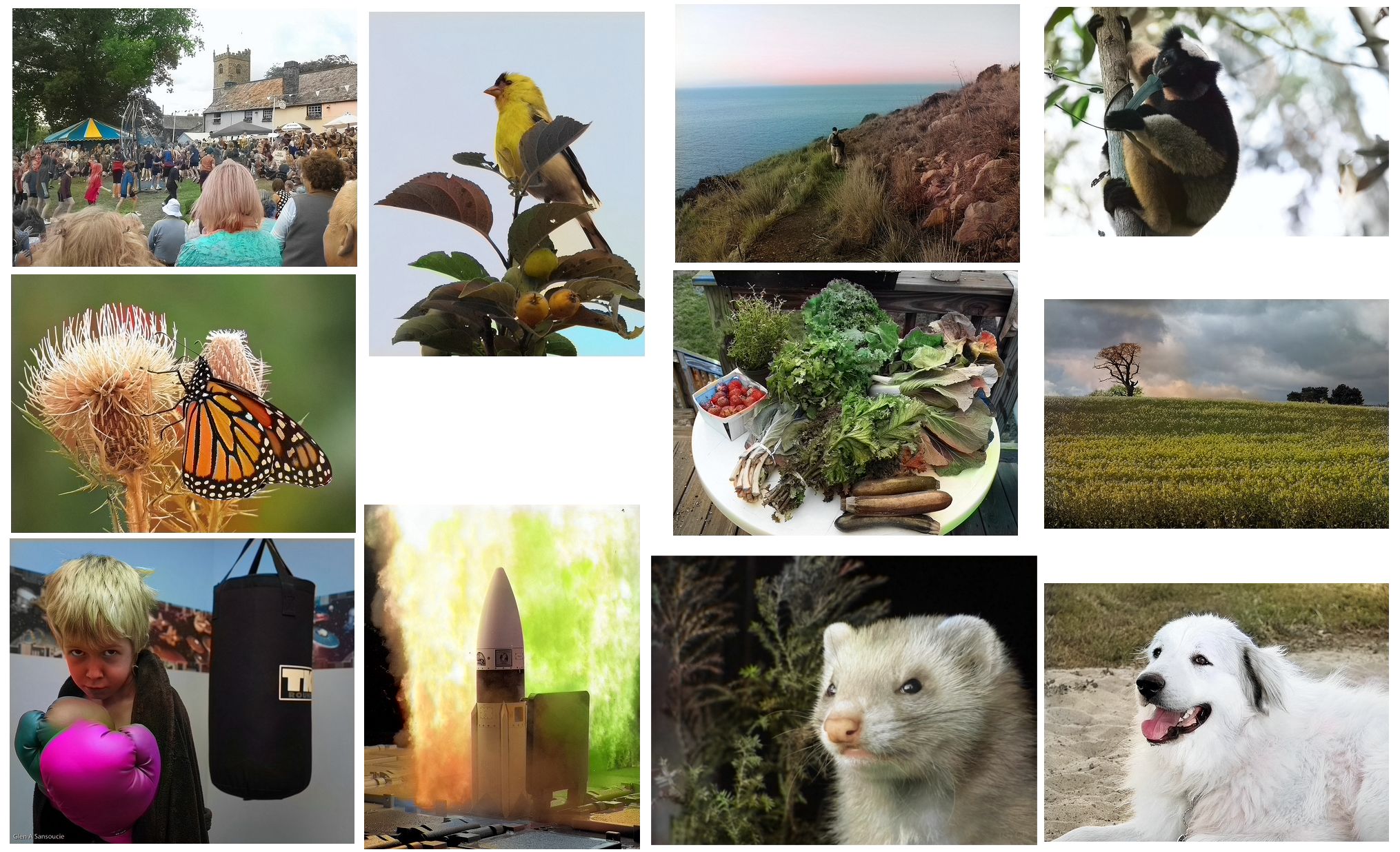}
  \caption{ Selected high resolution and non-square samples. }
  \label{fig:nonsquare}
\end{figure}

\setlength{\LTpre}{0pt}
\def \w { .135\linewidth }
\begin{longtable}{@{\hspace{.1cm}}c@{\hspace{.1cm}}c@{\hspace{.1cm}}c@{\hspace{.1cm}}c@{\hspace{.1cm}}c@{\hspace{.1cm}}c@{\hspace{.1cm}}c}
\caption{
Qualitative side by side comparison of colorizations produced by
various methods (LTBC: \cite{IizukaSIGGRAPH2016},
pix2pix: \cite{pix2pix2016},
VAE: \cite{diversecolorization2016},
LRAC: \cite{larsson2016learning},
CIC: \cite{zhang2016colorful},
\shortname: this paper,
G. Truth: original color).
These images are {\it randomly sampled} from the ImageNet test set.
}\\
\label{tab:SxS}

LTBC & pix2pix & cVAE & LRAC & CIC & \shortname & G. Truth \\
\endfirsthead
LTBC & pix2pix & cVAE & LRAC & CIC & \shortname & G. Truth \\
\endhead
{\includegraphics[width=\w]{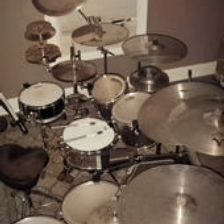}}
 & {\includegraphics[width=\w]{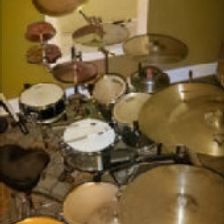}}
 & {\includegraphics[width=\w]{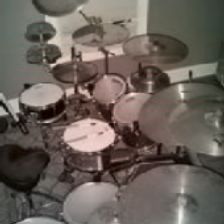}}
 & {\includegraphics[width=\w]{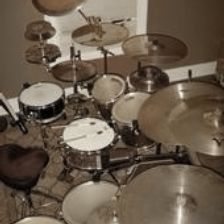}}
 & {\includegraphics[width=\w]{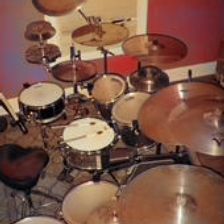}}
 & {\includegraphics[width=\w]{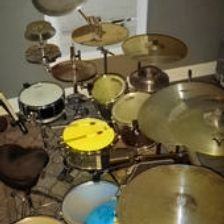}}
 & {\includegraphics[width=\w]{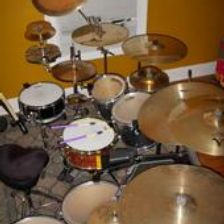}}
 \\
{\includegraphics[width=\w]{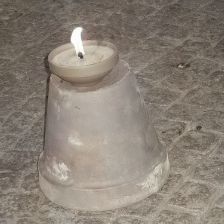}}
 & {\includegraphics[width=\w]{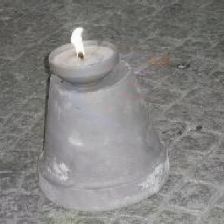}}
 & {\includegraphics[width=\w]{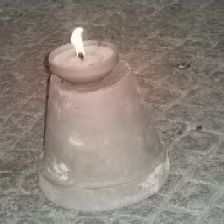}}
 & {\includegraphics[width=\w]{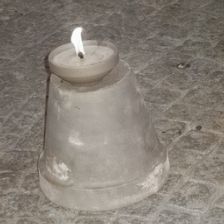}}
 & {\includegraphics[width=\w]{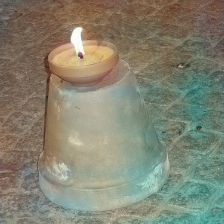}}
 & {\includegraphics[width=\w]{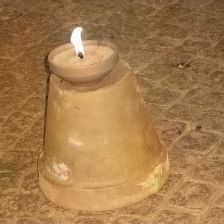}}
 & {\includegraphics[width=\w]{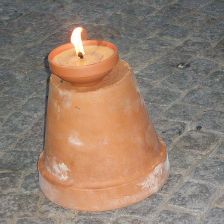}}
 \\
{\includegraphics[width=\w]{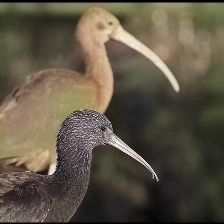}}
 & {\includegraphics[width=\w]{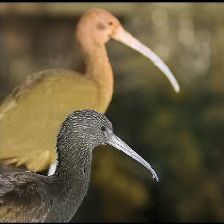}}
 & {\includegraphics[width=\w]{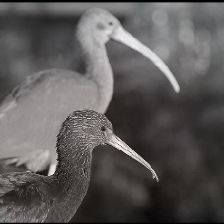}}
 & {\includegraphics[width=\w]{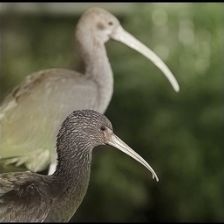}}
 & {\includegraphics[width=\w]{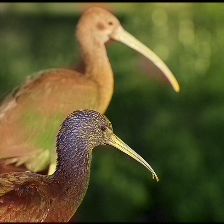}}
 & {\includegraphics[width=\w]{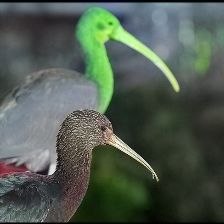}}
 & {\includegraphics[width=\w]{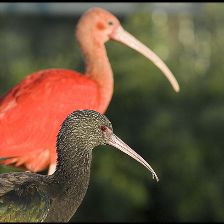}}
 \\
{\includegraphics[width=\w]{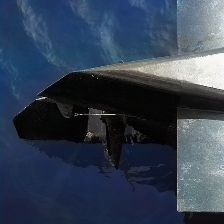}}
 & {\includegraphics[width=\w]{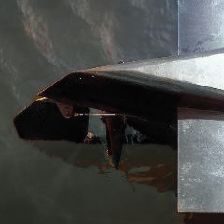}}
 & {\includegraphics[width=\w]{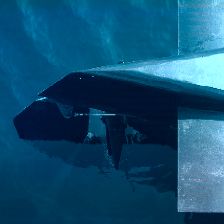}}
 & {\includegraphics[width=\w]{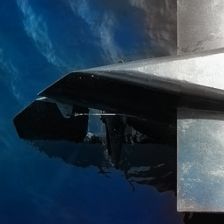}}
 & {\includegraphics[width=\w]{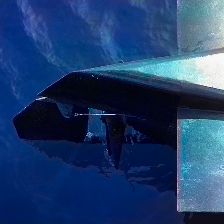}}
 & {\includegraphics[width=\w]{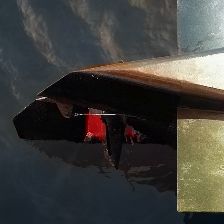}}
 & {\includegraphics[width=\w]{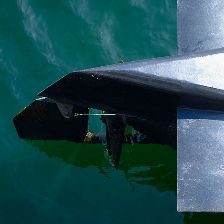}}
 \\
{\includegraphics[width=\w]{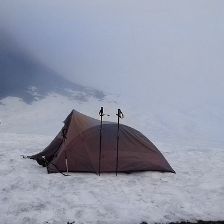}}
 & {\includegraphics[width=\w]{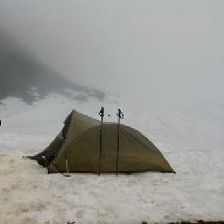}}
 & {\includegraphics[width=\w]{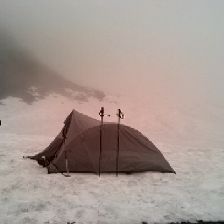}}
 & {\includegraphics[width=\w]{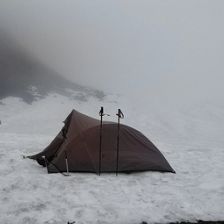}}
 & {\includegraphics[width=\w]{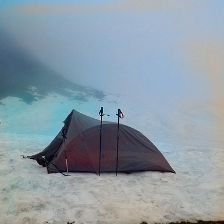}}
 & {\includegraphics[width=\w]{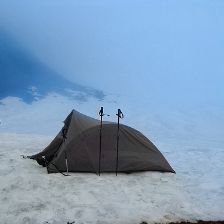}}
 & {\includegraphics[width=\w]{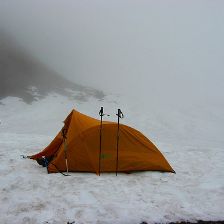}}
 \\
{\includegraphics[width=\w]{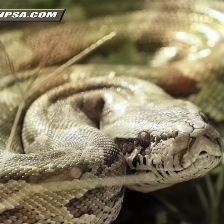}}
 & {\includegraphics[width=\w]{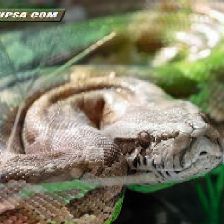}}
 & {\includegraphics[width=\w]{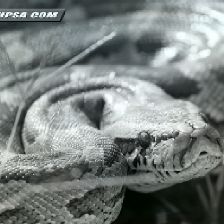}}
 & {\includegraphics[width=\w]{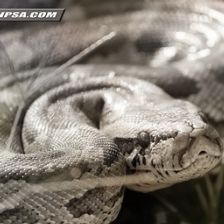}}
 & {\includegraphics[width=\w]{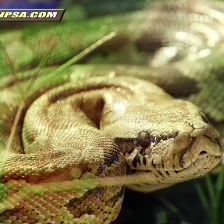}}
 & {\includegraphics[width=\w]{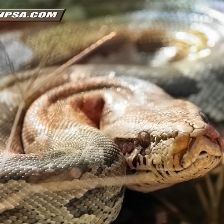}}
 & {\includegraphics[width=\w]{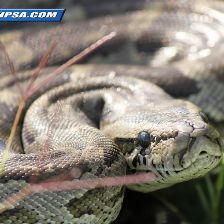}}
 \\
{\includegraphics[width=\w]{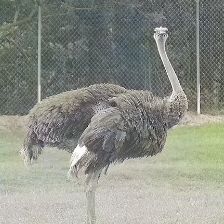}}
 & {\includegraphics[width=\w]{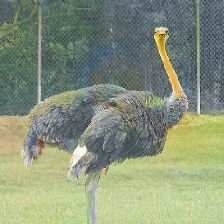}}
 & {\includegraphics[width=\w]{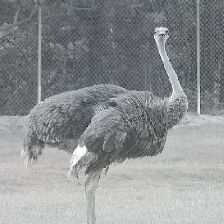}}
 & {\includegraphics[width=\w]{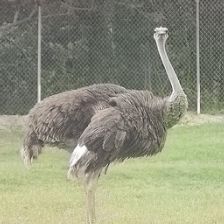}}
 & {\includegraphics[width=\w]{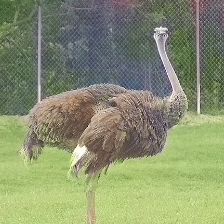}}
 & {\includegraphics[width=\w]{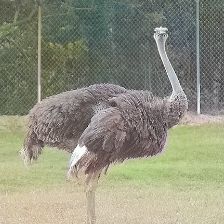}}
 & {\includegraphics[width=\w]{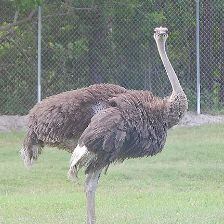}}
 \\
{\includegraphics[width=\w]{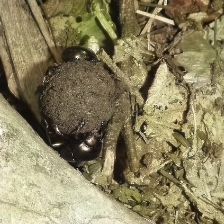}}
 & {\includegraphics[width=\w]{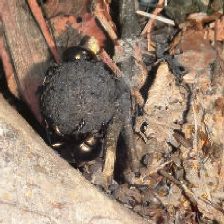}}
 & {\includegraphics[width=\w]{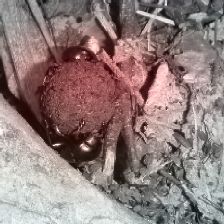}}
 & {\includegraphics[width=\w]{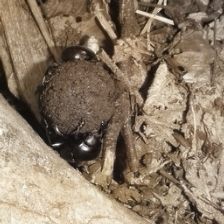}}
 & {\includegraphics[width=\w]{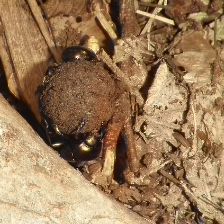}}
 & {\includegraphics[width=\w]{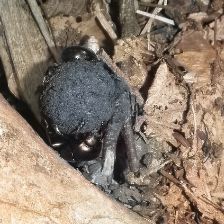}}
 & {\includegraphics[width=\w]{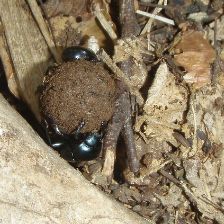}}
 \\
{\includegraphics[width=\w]{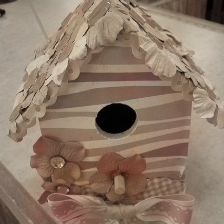}}
 & {\includegraphics[width=\w]{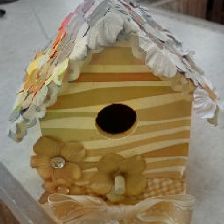}}
 & {\includegraphics[width=\w]{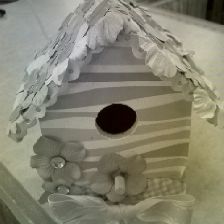}}
 & {\includegraphics[width=\w]{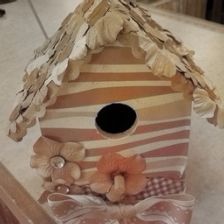}}
 & {\includegraphics[width=\w]{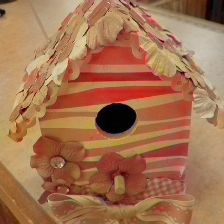}}
 & {\includegraphics[width=\w]{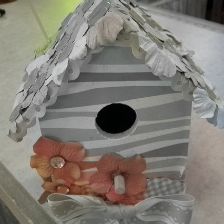}}
 & {\includegraphics[width=\w]{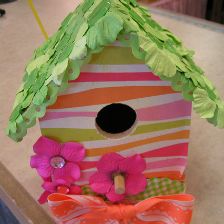}}
 \\
{\includegraphics[width=\w]{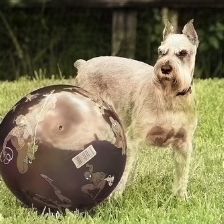}}
 & {\includegraphics[width=\w]{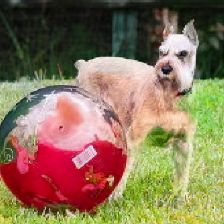}}
 & {\includegraphics[width=\w]{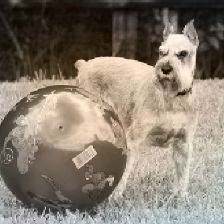}}
 & {\includegraphics[width=\w]{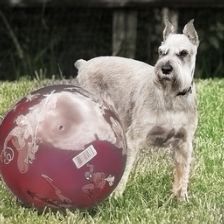}}
 & {\includegraphics[width=\w]{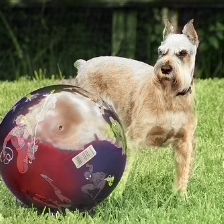}}
 & {\includegraphics[width=\w]{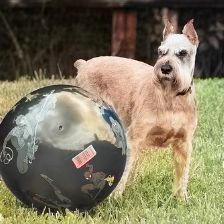}}
 & {\includegraphics[width=\w]{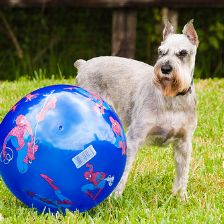}}
 \\
{\includegraphics[width=\w]{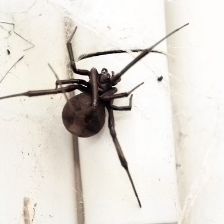}}
 & {\includegraphics[width=\w]{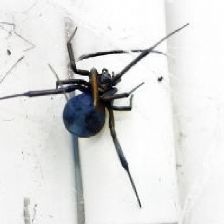}}
 & {\includegraphics[width=\w]{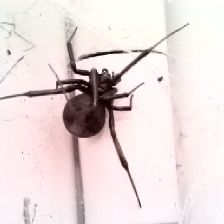}}
 & {\includegraphics[width=\w]{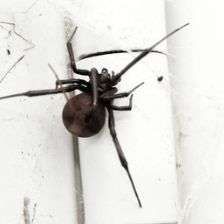}}
 & {\includegraphics[width=\w]{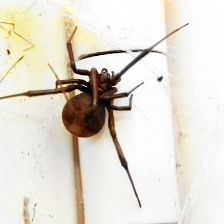}}
 & {\includegraphics[width=\w]{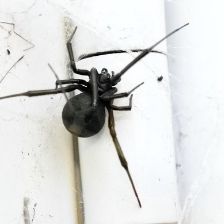}}
 & {\includegraphics[width=\w]{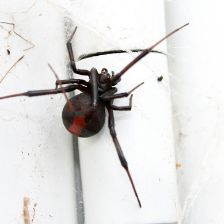}}
 \\
{\includegraphics[width=\w]{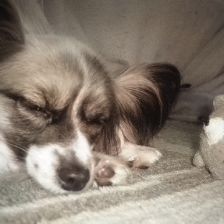}}
 & {\includegraphics[width=\w]{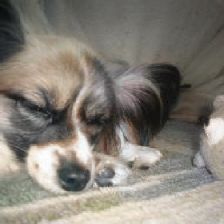}}
 & {\includegraphics[width=\w]{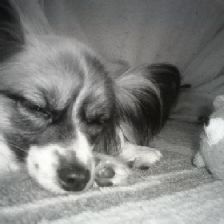}}
 & {\includegraphics[width=\w]{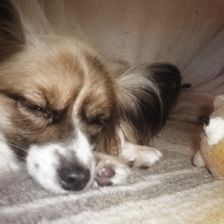}}
 & {\includegraphics[width=\w]{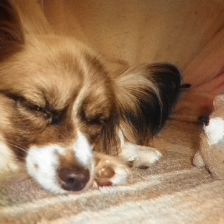}}
 & {\includegraphics[width=\w]{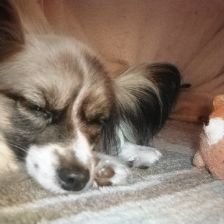}}
 & {\includegraphics[width=\w]{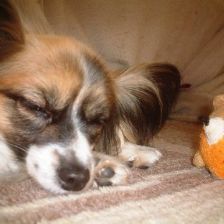}}
 \\
{\includegraphics[width=\w]{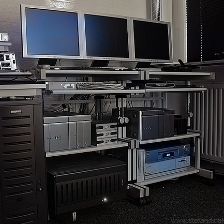}}
 & {\includegraphics[width=\w]{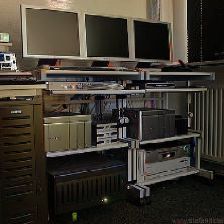}}
 & {\includegraphics[width=\w]{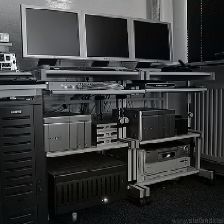}}
 & {\includegraphics[width=\w]{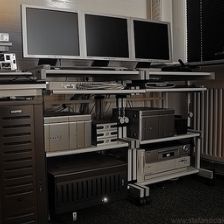}}
 & {\includegraphics[width=\w]{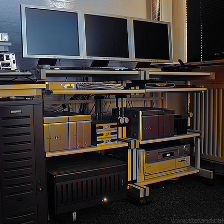}}
 & {\includegraphics[width=\w]{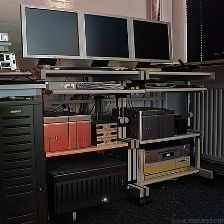}}
 & {\includegraphics[width=\w]{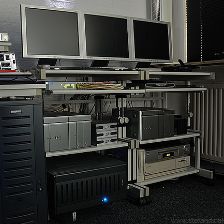}}
 \\
{\includegraphics[width=\w]{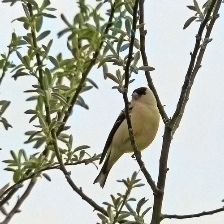}}
 & {\includegraphics[width=\w]{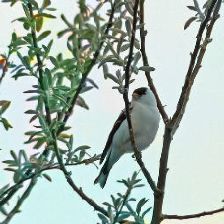}}
 & {\includegraphics[width=\w]{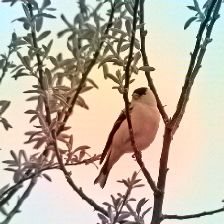}}
 & {\includegraphics[width=\w]{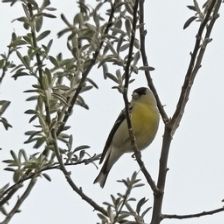}}
 & {\includegraphics[width=\w]{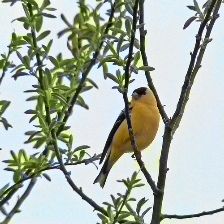}}
 & {\includegraphics[width=\w]{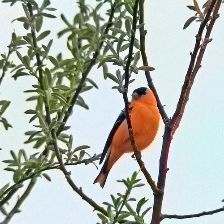}}
 & {\includegraphics[width=\w]{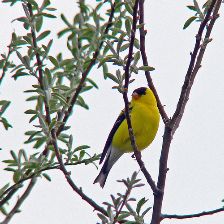}}
 \\
{\includegraphics[width=\w]{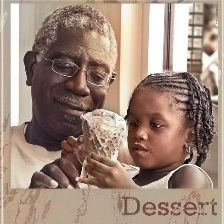}}
 & {\includegraphics[width=\w]{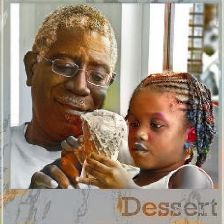}}
 & {\includegraphics[width=\w]{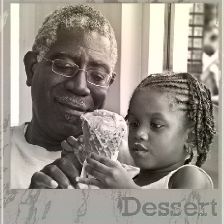}}
 & {\includegraphics[width=\w]{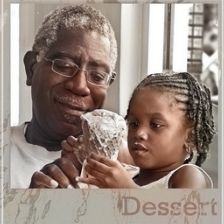}}
 & {\includegraphics[width=\w]{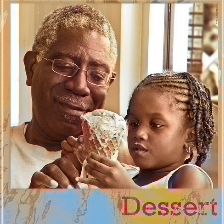}}
 & {\includegraphics[width=\w]{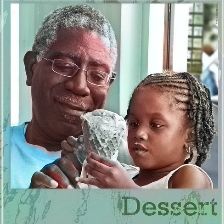}}
 & {\includegraphics[width=\w]{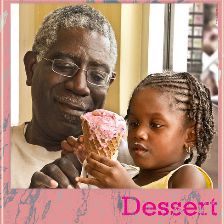}}
 \\
{\includegraphics[width=\w]{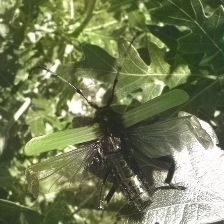}}
 & {\includegraphics[width=\w]{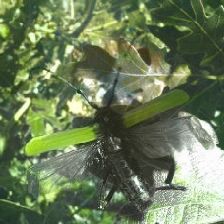}}
 & {\includegraphics[width=\w]{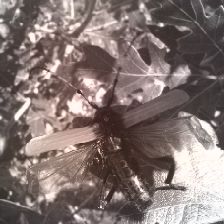}}
 & {\includegraphics[width=\w]{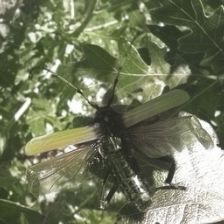}}
 & {\includegraphics[width=\w]{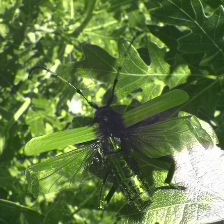}}
 & {\includegraphics[width=\w]{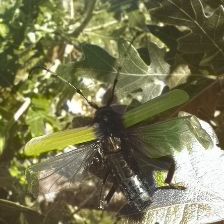}}
 & {\includegraphics[width=\w]{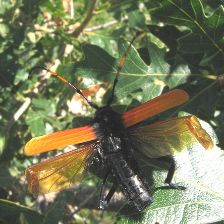}}
 \\
{\includegraphics[width=\w]{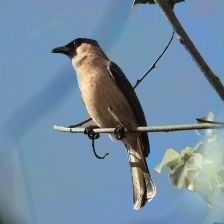}}
 & {\includegraphics[width=\w]{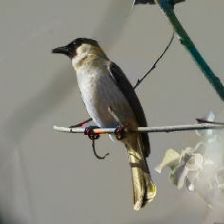}}
 & {\includegraphics[width=\w]{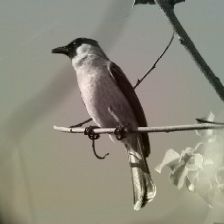}}
 & {\includegraphics[width=\w]{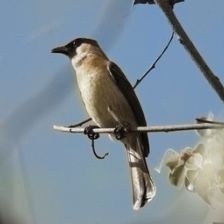}}
 & {\includegraphics[width=\w]{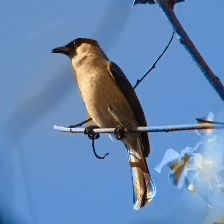}}
 & {\includegraphics[width=\w]{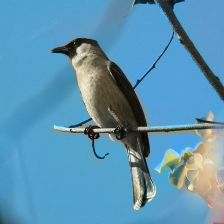}}
 & {\includegraphics[width=\w]{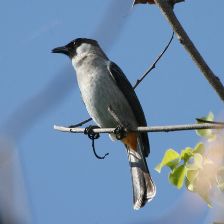}}
 \\
{\includegraphics[width=\w]{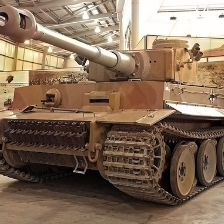}}
 & {\includegraphics[width=\w]{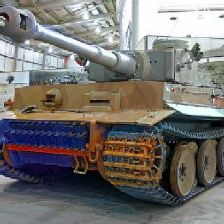}}
 & {\includegraphics[width=\w]{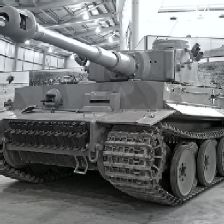}}
 & {\includegraphics[width=\w]{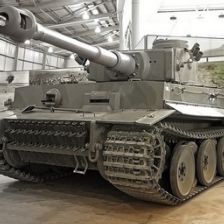}}
 & {\includegraphics[width=\w]{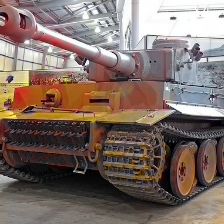}}
 & {\includegraphics[width=\w]{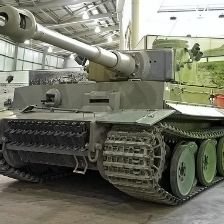}}
 & {\includegraphics[width=\w]{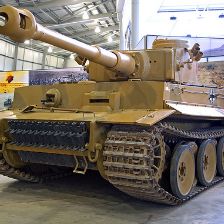}}
 \\
{\includegraphics[width=\w]{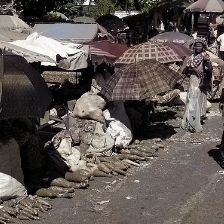}}
 & {\includegraphics[width=\w]{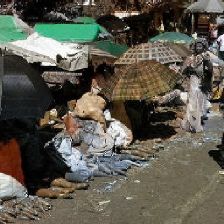}}
 & {\includegraphics[width=\w]{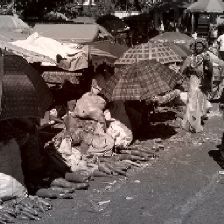}}
 & {\includegraphics[width=\w]{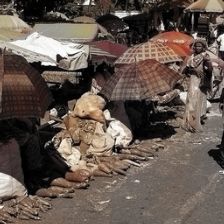}}
 & {\includegraphics[width=\w]{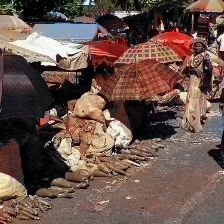}}
 & {\includegraphics[width=\w]{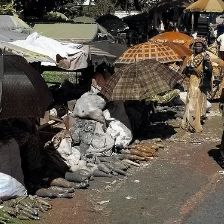}}
 & {\includegraphics[width=\w]{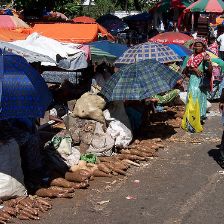}}
 \\
{\includegraphics[width=\w]{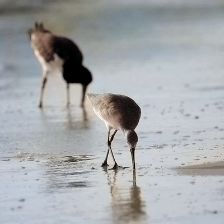}}
 & {\includegraphics[width=\w]{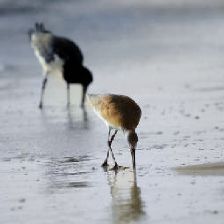}}
 & {\includegraphics[width=\w]{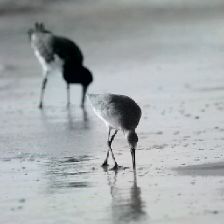}}
 & {\includegraphics[width=\w]{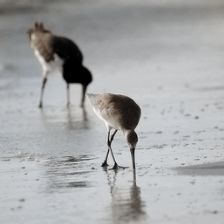}}
 & {\includegraphics[width=\w]{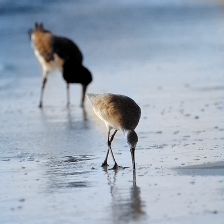}}
 & {\includegraphics[width=\w]{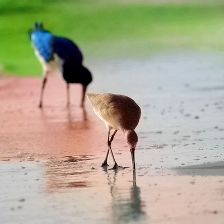}}
 & {\includegraphics[width=\w]{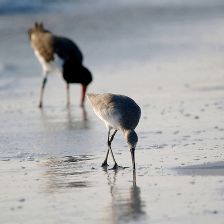}}
 \\
{\includegraphics[width=\w]{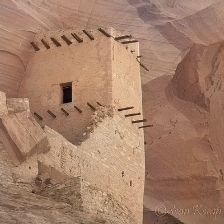}}
 & {\includegraphics[width=\w]{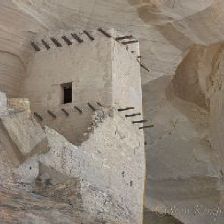}}
 & {\includegraphics[width=\w]{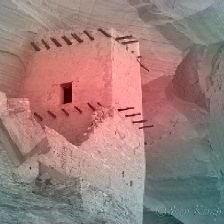}}
 & {\includegraphics[width=\w]{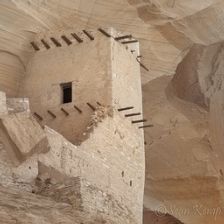}}
 & {\includegraphics[width=\w]{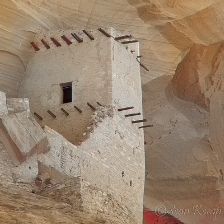}}
 & {\includegraphics[width=\w]{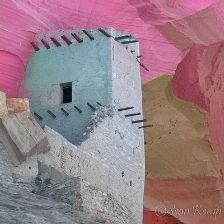}}
 & {\includegraphics[width=\w]{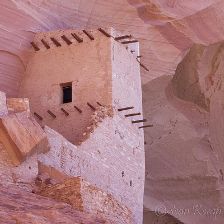}}
 \\
{\includegraphics[width=\w]{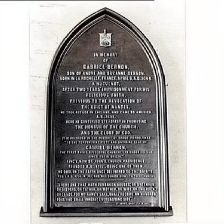}}
 & {\includegraphics[width=\w]{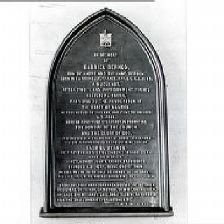}}
 & {\includegraphics[width=\w]{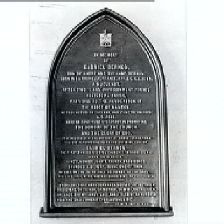}}
 & {\includegraphics[width=\w]{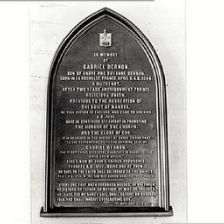}}
 & {\includegraphics[width=\w]{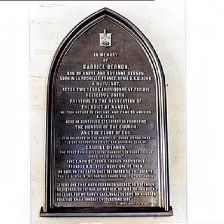}}
 & {\includegraphics[width=\w]{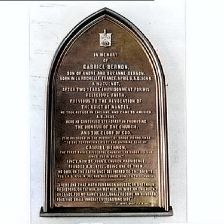}}
 & {\includegraphics[width=\w]{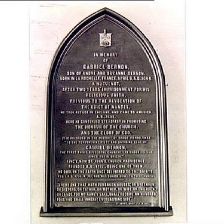}}
 \\
{\includegraphics[width=\w]{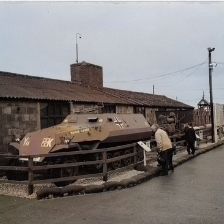}}
 & {\includegraphics[width=\w]{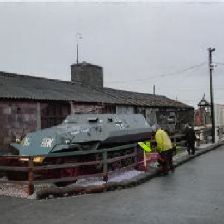}}
 & {\includegraphics[width=\w]{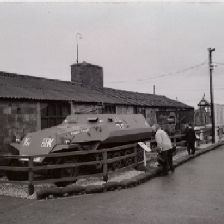}}
 & {\includegraphics[width=\w]{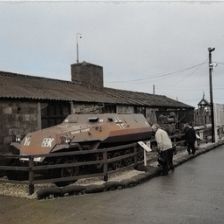}}
 & {\includegraphics[width=\w]{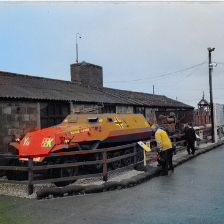}}
 & {\includegraphics[width=\w]{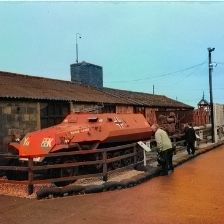}}
 & {\includegraphics[width=\w]{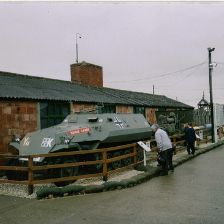}}
 \\
{\includegraphics[width=\w]{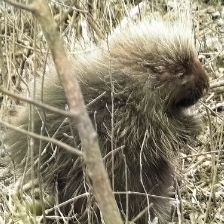}}
 & {\includegraphics[width=\w]{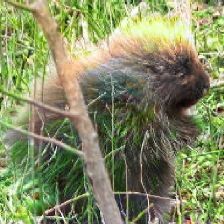}}
 & {\includegraphics[width=\w]{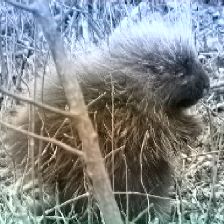}}
 & {\includegraphics[width=\w]{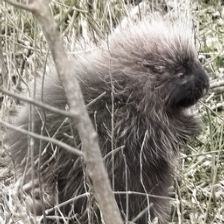}}
 & {\includegraphics[width=\w]{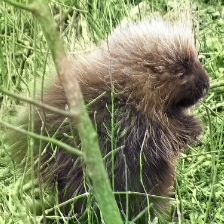}}
 & {\includegraphics[width=\w]{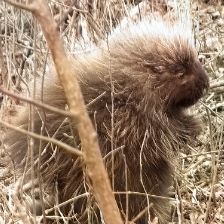}}
 & {\includegraphics[width=\w]{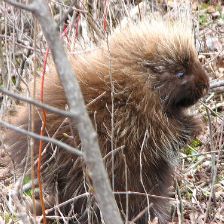}}
 \\
{\includegraphics[width=\w]{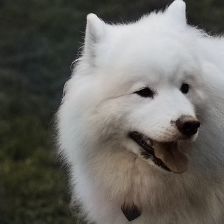}}
 & {\includegraphics[width=\w]{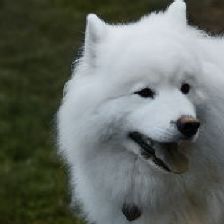}}
 & {\includegraphics[width=\w]{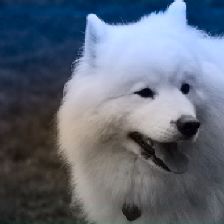}}
 & {\includegraphics[width=\w]{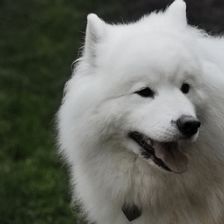}}
 & {\includegraphics[width=\w]{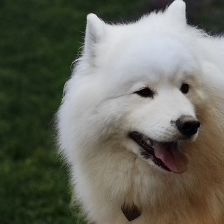}}
 & {\includegraphics[width=\w]{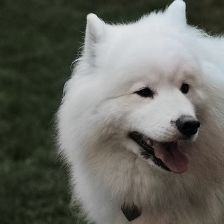}}
 & {\includegraphics[width=\w]{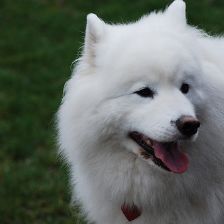}}
 \\
{\includegraphics[width=\w]{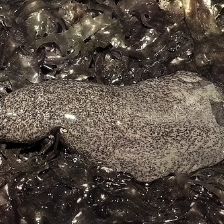}}
 & {\includegraphics[width=\w]{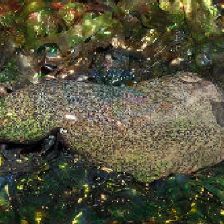}}
 & {\includegraphics[width=\w]{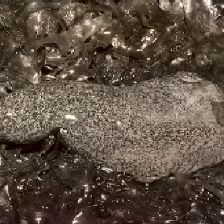}}
 & {\includegraphics[width=\w]{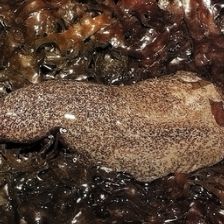}}
 & {\includegraphics[width=\w]{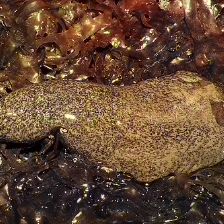}}
 & {\includegraphics[width=\w]{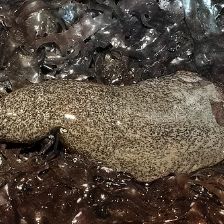}}
 & {\includegraphics[width=\w]{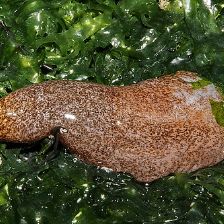}}
 \\
{\includegraphics[width=\w]{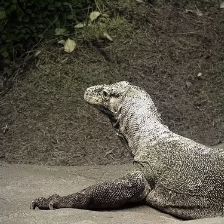}}
 & {\includegraphics[width=\w]{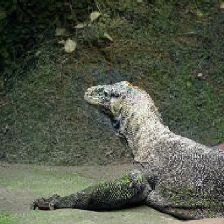}}
 & {\includegraphics[width=\w]{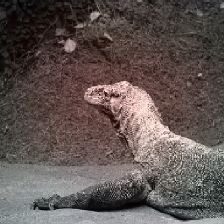}}
 & {\includegraphics[width=\w]{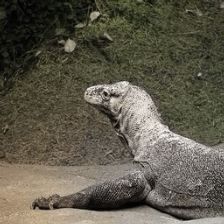}}
 & {\includegraphics[width=\w]{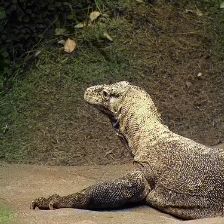}}
 & {\includegraphics[width=\w]{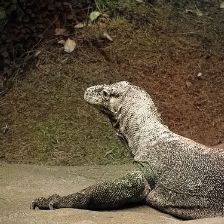}}
 & {\includegraphics[width=\w]{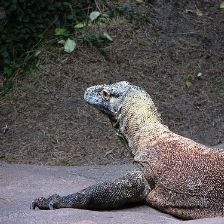}}
 \\
{\includegraphics[width=\w]{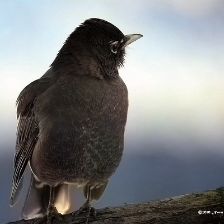}}
 & {\includegraphics[width=\w]{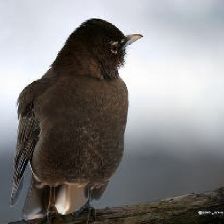}}
 & {\includegraphics[width=\w]{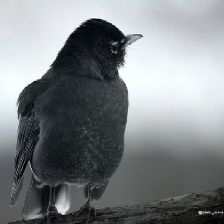}}
 & {\includegraphics[width=\w]{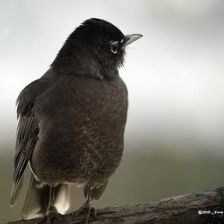}}
 & {\includegraphics[width=\w]{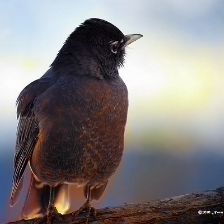}}
 & {\includegraphics[width=\w]{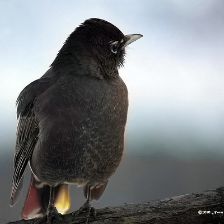}}
 & {\includegraphics[width=\w]{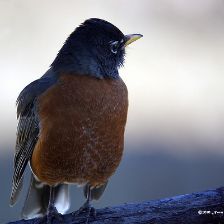}}
 \\
{\includegraphics[width=\w]{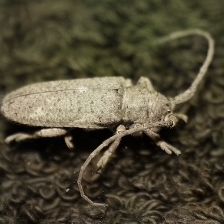}}
 & {\includegraphics[width=\w]{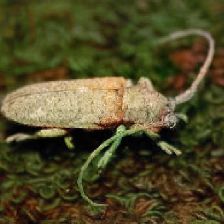}}
 & {\includegraphics[width=\w]{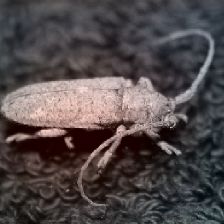}}
 & {\includegraphics[width=\w]{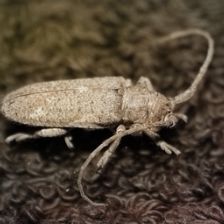}}
 & {\includegraphics[width=\w]{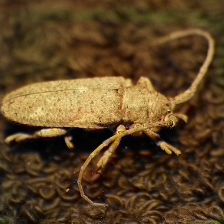}}
 & {\includegraphics[width=\w]{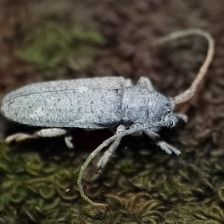}}
 & {\includegraphics[width=\w]{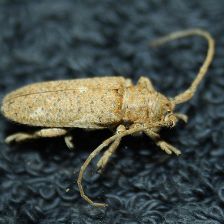}}
 \\
{\includegraphics[width=\w]{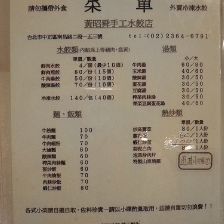}}
 & {\includegraphics[width=\w]{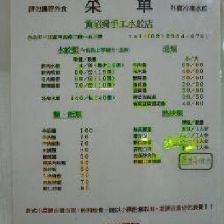}}
 & {\includegraphics[width=\w]{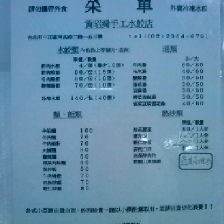}}
 & {\includegraphics[width=\w]{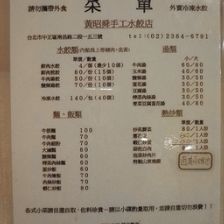}}
 & {\includegraphics[width=\w]{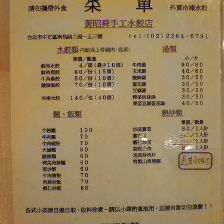}}
 & {\includegraphics[width=\w]{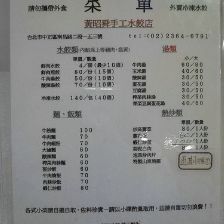}}
 & {\includegraphics[width=\w]{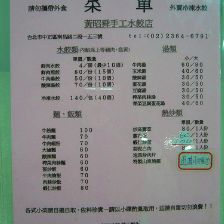}}
 \\
{\includegraphics[width=\w]{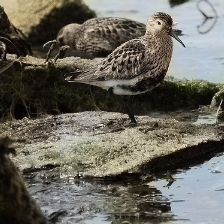}}
 & {\includegraphics[width=\w]{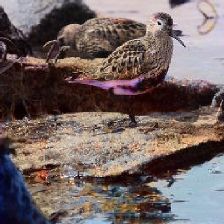}}
 & {\includegraphics[width=\w]{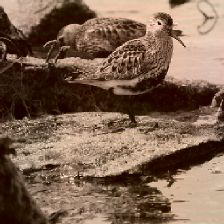}}
 & {\includegraphics[width=\w]{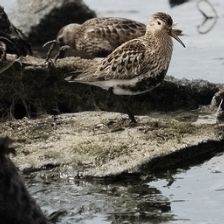}}
 & {\includegraphics[width=\w]{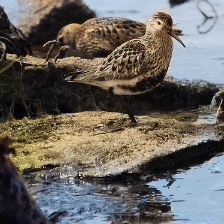}}
 & {\includegraphics[width=\w]{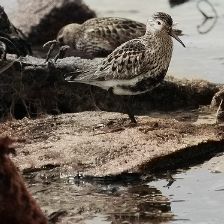}}
 & {\includegraphics[width=\w]{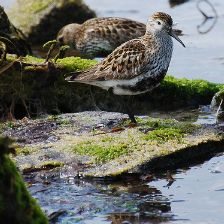}}
 \\
{\includegraphics[width=\w]{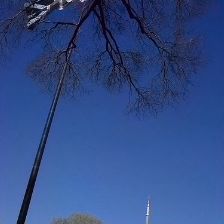}}
 & {\includegraphics[width=\w]{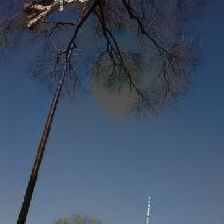}}
 & {\includegraphics[width=\w]{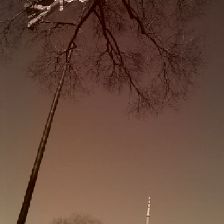}}
 & {\includegraphics[width=\w]{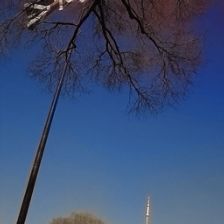}}
 & {\includegraphics[width=\w]{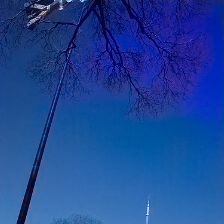}}
 & {\includegraphics[width=\w]{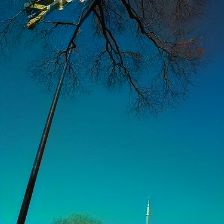}}
 & {\includegraphics[width=\w]{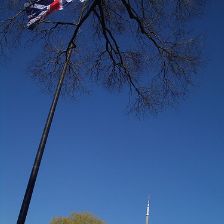}}
 \\
{\includegraphics[width=\w]{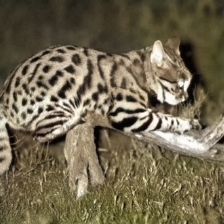}}
 & {\includegraphics[width=\w]{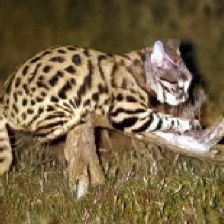}}
 & {\includegraphics[width=\w]{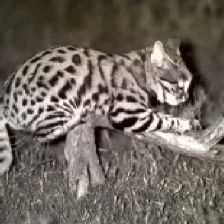}}
 & {\includegraphics[width=\w]{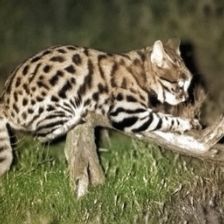}}
 & {\includegraphics[width=\w]{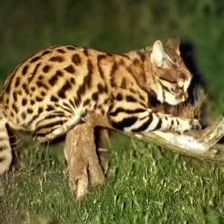}}
 & {\includegraphics[width=\w]{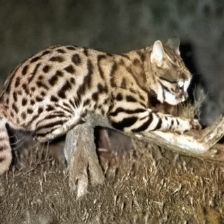}}
 & {\includegraphics[width=\w]{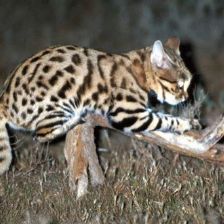}}
 \\
{\includegraphics[width=\w]{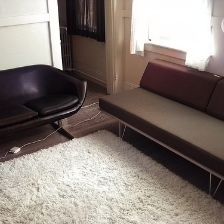}}
 & {\includegraphics[width=\w]{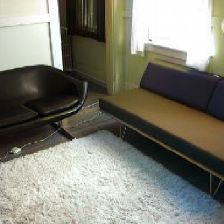}}
 & {\includegraphics[width=\w]{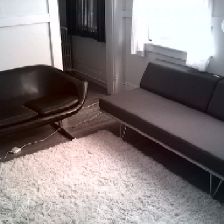}}
 & {\includegraphics[width=\w]{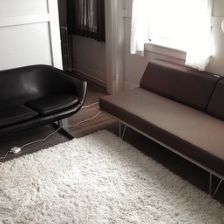}}
 & {\includegraphics[width=\w]{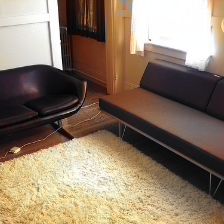}}
 & {\includegraphics[width=\w]{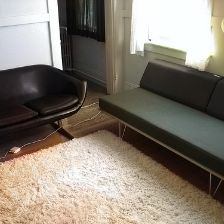}}
 & {\includegraphics[width=\w]{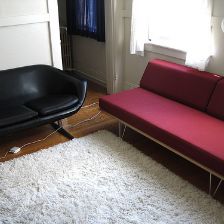}}
 \\
{\includegraphics[width=\w]{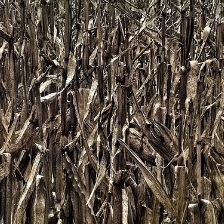}}
 & {\includegraphics[width=\w]{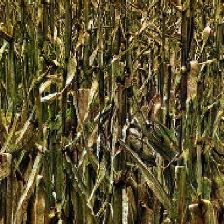}}
 & {\includegraphics[width=\w]{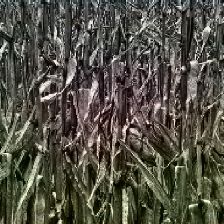}}
 & {\includegraphics[width=\w]{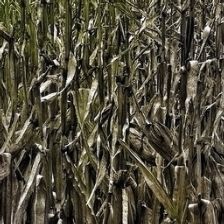}}
 & {\includegraphics[width=\w]{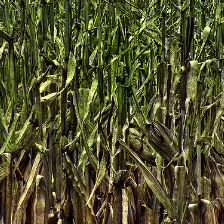}}
 & {\includegraphics[width=\w]{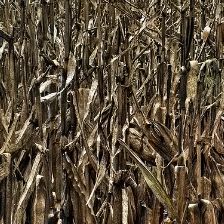}}
 & {\includegraphics[width=\w]{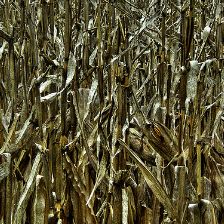}}
 \\
{\includegraphics[width=\w]{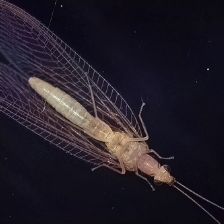}}
 & {\includegraphics[width=\w]{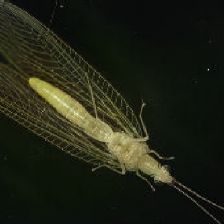}}
 & {\includegraphics[width=\w]{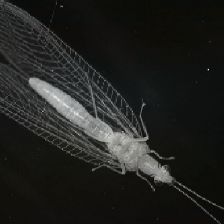}}
 & {\includegraphics[width=\w]{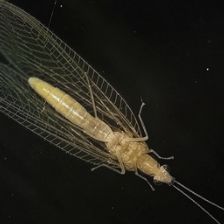}}
 & {\includegraphics[width=\w]{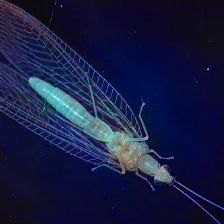}}
 & {\includegraphics[width=\w]{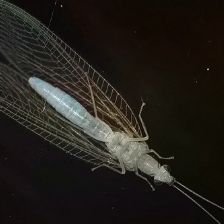}}
 & {\includegraphics[width=\w]{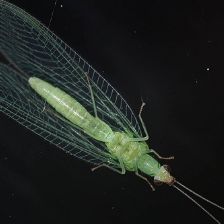}}
 \\
{\includegraphics[width=\w]{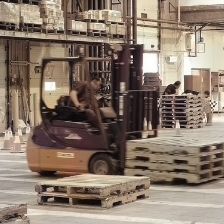}}
 & {\includegraphics[width=\w]{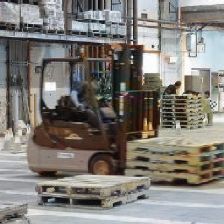}}
 & {\includegraphics[width=\w]{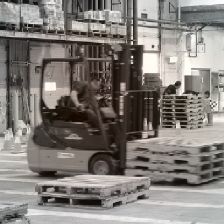}}
 & {\includegraphics[width=\w]{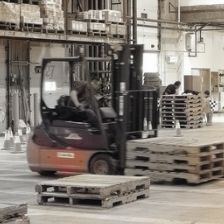}}
 & {\includegraphics[width=\w]{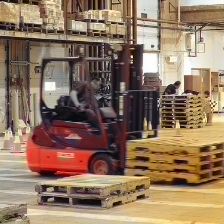}}
 & {\includegraphics[width=\w]{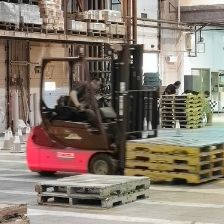}}
 & {\includegraphics[width=\w]{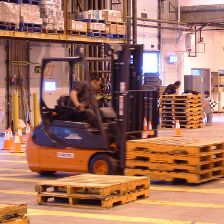}}
 \\
{\includegraphics[width=\w]{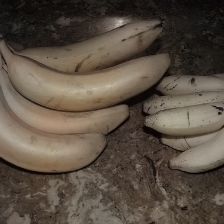}}
 & {\includegraphics[width=\w]{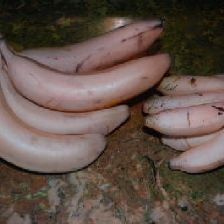}}
 & {\includegraphics[width=\w]{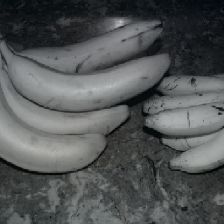}}
 & {\includegraphics[width=\w]{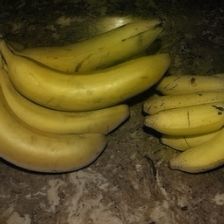}}
 & {\includegraphics[width=\w]{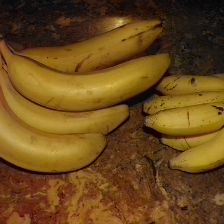}}
 & {\includegraphics[width=\w]{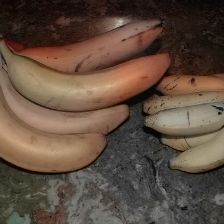}}
 & {\includegraphics[width=\w]{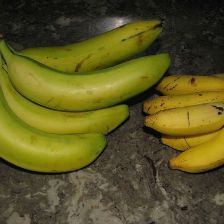}}
 \\
{\includegraphics[width=\w]{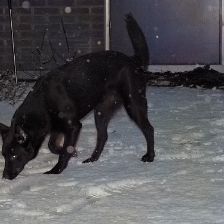}}
 & {\includegraphics[width=\w]{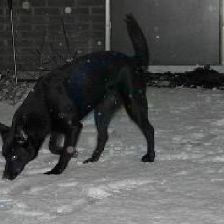}}
 & {\includegraphics[width=\w]{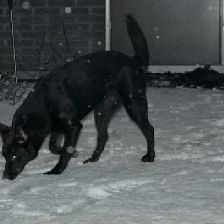}}
 & {\includegraphics[width=\w]{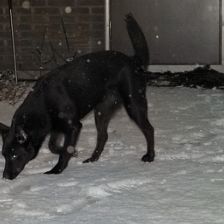}}
 & {\includegraphics[width=\w]{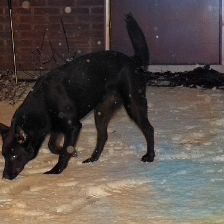}}
 & {\includegraphics[width=\w]{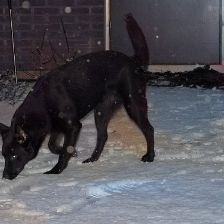}}
 & {\includegraphics[width=\w]{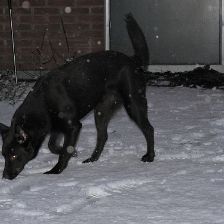}}
 \\
{\includegraphics[width=\w]{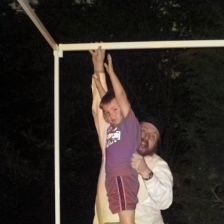}}
 & {\includegraphics[width=\w]{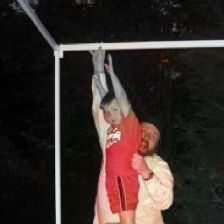}}
 & {\includegraphics[width=\w]{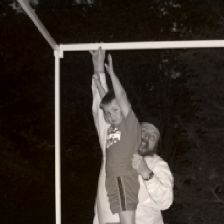}}
 & {\includegraphics[width=\w]{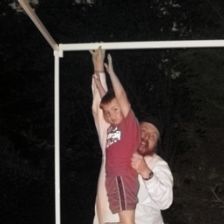}}
 & {\includegraphics[width=\w]{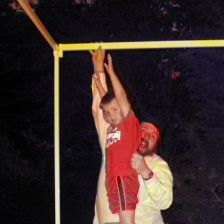}}
 & {\includegraphics[width=\w]{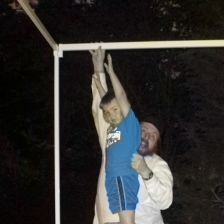}}
 & {\includegraphics[width=\w]{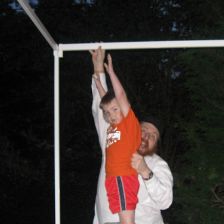}}
 \\
{\includegraphics[width=\w]{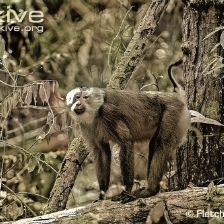}}
 & {\includegraphics[width=\w]{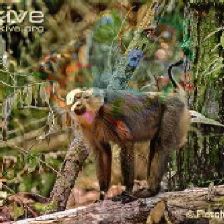}}
 & {\includegraphics[width=\w]{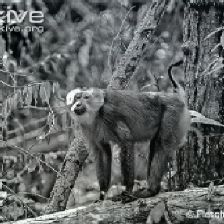}}
 & {\includegraphics[width=\w]{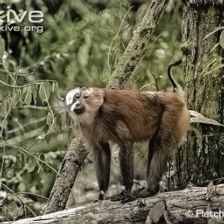}}
 & {\includegraphics[width=\w]{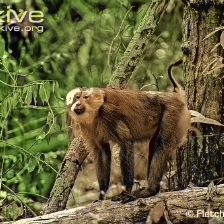}}
 & {\includegraphics[width=\w]{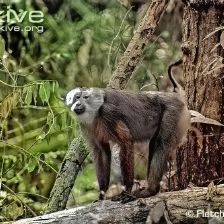}}
 & {\includegraphics[width=\w]{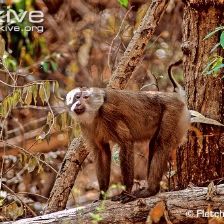}}
 \\
{\includegraphics[width=\w]{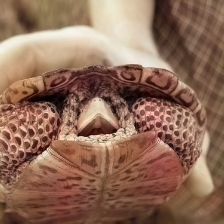}}
 & {\includegraphics[width=\w]{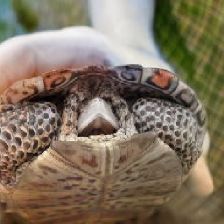}}
 & {\includegraphics[width=\w]{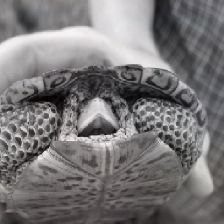}}
 & {\includegraphics[width=\w]{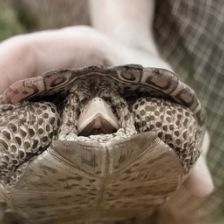}}
 & {\includegraphics[width=\w]{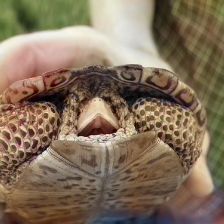}}
 & {\includegraphics[width=\w]{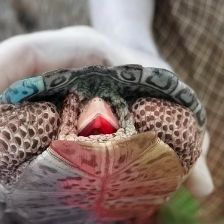}}
 & {\includegraphics[width=\w]{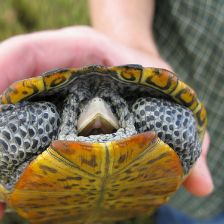}}
 \\
{\includegraphics[width=\w]{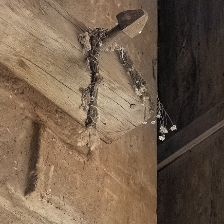}}
 & {\includegraphics[width=\w]{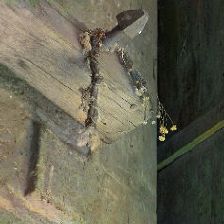}}
 & {\includegraphics[width=\w]{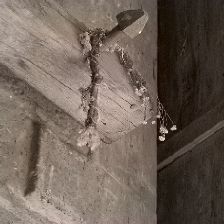}}
 & {\includegraphics[width=\w]{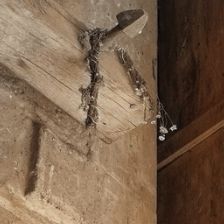}}
 & {\includegraphics[width=\w]{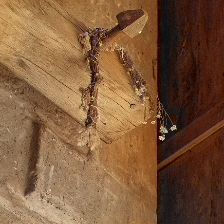}}
 & {\includegraphics[width=\w]{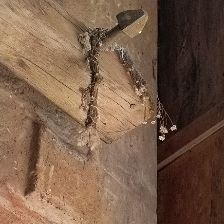}}
 & {\includegraphics[width=\w]{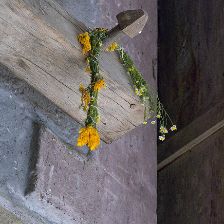}}
 \\
{\includegraphics[width=\w]{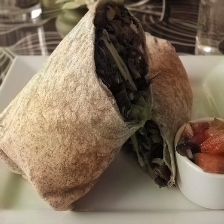}}
 & {\includegraphics[width=\w]{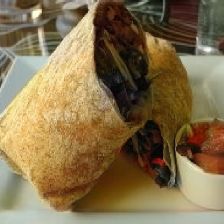}}
 & {\includegraphics[width=\w]{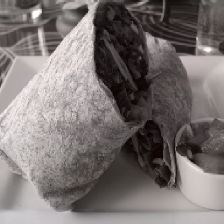}}
 & {\includegraphics[width=\w]{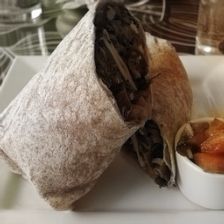}}
 & {\includegraphics[width=\w]{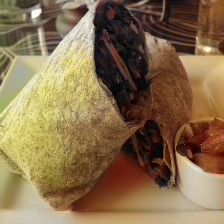}}
 & {\includegraphics[width=\w]{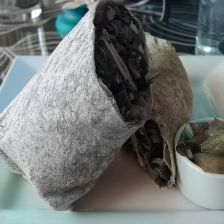}}
 & {\includegraphics[width=\w]{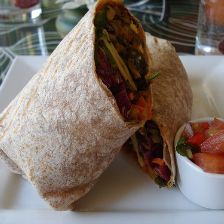}}
 \\
{\includegraphics[width=\w]{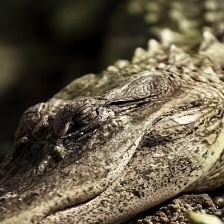}}
 & {\includegraphics[width=\w]{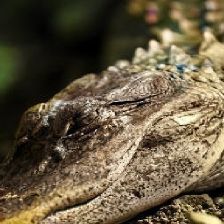}}
 & {\includegraphics[width=\w]{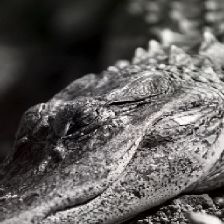}}
 & {\includegraphics[width=\w]{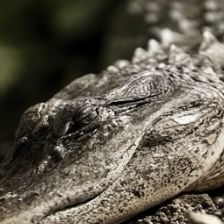}}
 & {\includegraphics[width=\w]{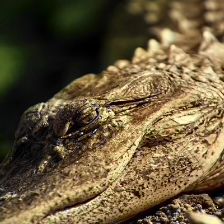}}
 & {\includegraphics[width=\w]{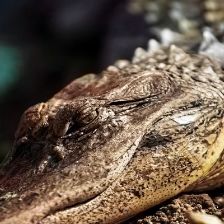}}
 & {\includegraphics[width=\w]{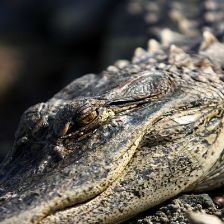}}
 \\
{\includegraphics[width=\w]{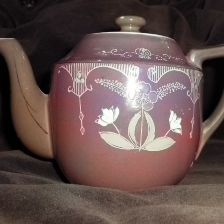}}
 & {\includegraphics[width=\w]{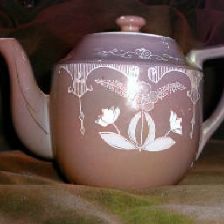}}
 & {\includegraphics[width=\w]{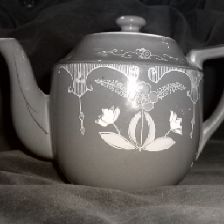}}
 & {\includegraphics[width=\w]{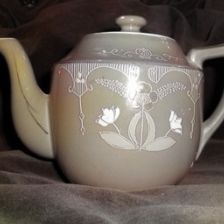}}
 & {\includegraphics[width=\w]{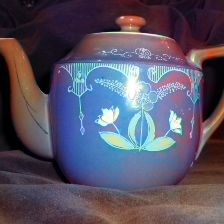}}
 & {\includegraphics[width=\w]{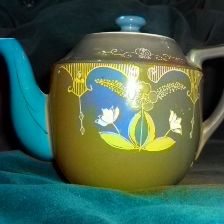}}
 & {\includegraphics[width=\w]{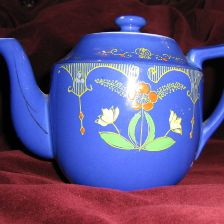}}
 \\
{\includegraphics[width=\w]{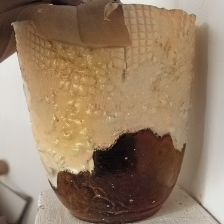}}
 & {\includegraphics[width=\w]{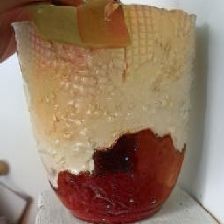}}
 & {\includegraphics[width=\w]{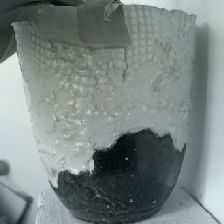}}
 & {\includegraphics[width=\w]{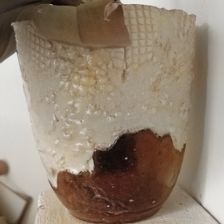}}
 & {\includegraphics[width=\w]{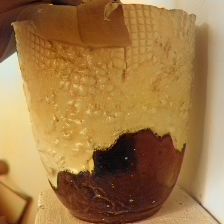}}
 & {\includegraphics[width=\w]{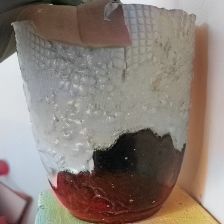}}
 & {\includegraphics[width=\w]{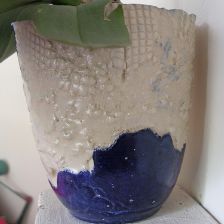}}
 \\
{\includegraphics[width=\w]{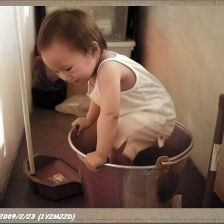}}
 & {\includegraphics[width=\w]{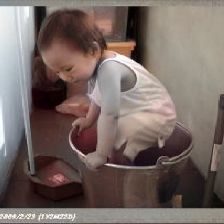}}
 & {\includegraphics[width=\w]{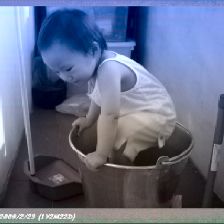}}
 & {\includegraphics[width=\w]{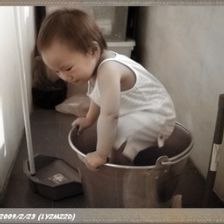}}
 & {\includegraphics[width=\w]{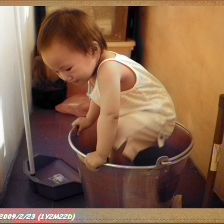}}
 & {\includegraphics[width=\w]{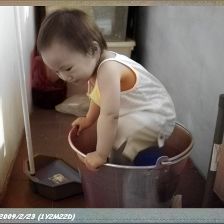}}
 & {\includegraphics[width=\w]{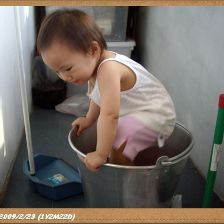}}
 \\
{\includegraphics[width=\w]{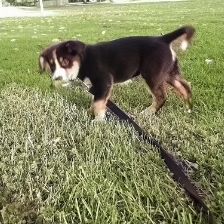}}
 & {\includegraphics[width=\w]{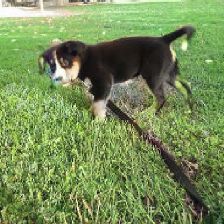}}
 & {\includegraphics[width=\w]{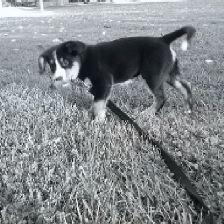}}
 & {\includegraphics[width=\w]{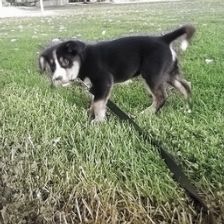}}
 & {\includegraphics[width=\w]{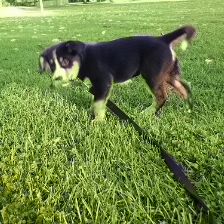}}
 & {\includegraphics[width=\w]{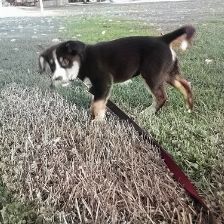}}
 & {\includegraphics[width=\w]{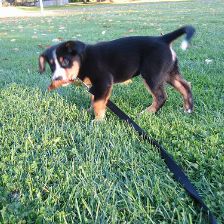}}


\end{longtable}

\end{document}